\documentclass[10pt,twocolumn,letterpaper]{article}

\usepackage{cvpr}              
\usepackage{amsthm}

\usepackage{booktabs} 
\usepackage{multirow}
\usepackage{makecell}
\usepackage{diagbox}   
\usepackage{bm}
\usepackage{amsmath}
\usepackage{amssymb}
\usepackage{subcaption} 
\usepackage{graphicx}

\newtheorem{lemma}{Lemma}
\newtheorem{theorem}{Theorem}
\newtheorem{definition}{Definition}

\definecolor{cvprblue}{rgb}{0.21,0.49,0.74}
\usepackage[pagebackref,breaklinks,colorlinks,allcolors=cvprblue]{hyperref}

\title{From Correlation to Causation: Max-Pooling-Based Multi-Instance Learning Leads to More Robust Whole Slide Image Classification}

\author{
Xin Liu\textsuperscript{\rm 1,2,3},
Weijia Zhang\textsuperscript{\rm 4},
Wei Tang\textsuperscript{\rm 2,3,5},
Thuc Duy Le\textsuperscript{\rm 1},
Jiuyong Li\textsuperscript{\rm 1},
Lin Liu\textsuperscript{\rm 1},
Min-Ling Zhang\textsuperscript{\rm 2,3\thanks{Corresponding author.}}\\
\textsuperscript{\rm 1} UniSA STEM, University of South Australia, Adelaide, Australia\\
\textsuperscript{\rm 2} School of Computer Science and Engineering, Southeast University, Nanjing 210096, China\\
\textsuperscript{\rm 3} Key Lab. of Computer Network and Information Integration (Southeast University), MoE, China\\
\textsuperscript{\rm 4} School of Information and Physical Sciences, The University of Newcastle, NSW 2308, Australia\\
\textsuperscript{\rm 5} Mohamed bin Zayed University of Artificial Intelligence (MBZUAI), Abu Dhabi, UAE\\
{\tt\small xin.liu@mymail.unisa.edu.au, weijia.zhang@newcastle.edu.au}\\
{\tt\small \{tangw, zhangml\}@seu.edu.cn, \{thuc.le, jiuyong.li, lin.liu\}@unisa.edu.au}
}

\begin{document}
\maketitle
\AddToShipoutPicture*{\AtTextLowerLeft{\put(0,-28){\small Under review.}}}

\begin{abstract}

In whole slide images (WSIs) analysis, attention-based multi-instance learning (MIL) models are susceptible to spurious correlations and degrade under domain shift. These methods may assign high attention weights to non-tumor regions, such as staining biases or artifacts, leading to unreliable tumor region localization. In this paper, we revisit max-pooling-based MIL methods from a causal perspective. Under mild assumptions, our theoretical results demonstrate that max-pooling encourages the model to focus on causal factors while ignoring bias-related factors. Furthermore, we discover that existing max-pooling-based methods may overfit the training set through rote memorization of instance features and fail to learn meaningful patterns. To address these issues, we propose FocusMIL, which couples max-pooling with an \emph{instance-level variational information bottleneck} (VIB) to learn compact, predictive latent representations, and employs a \emph{multi-bag mini-batch} scheme to stabilize optimization. We conduct comprehensive experiments on three real-world datasets and one semi-synthetic dataset. The results show that, by capturing causal factors, FocusMIL exhibits significant advantages in out-of-distribution scenarios and instance-level tumor region localization tasks.

\end{abstract}

\section{Introduction}

Whole slide images (WSIs) analysis is widely used for computer-aided tumour diagnosis and prognosis \cite{lu2021clam,yao2020deepattenMISL,chen2021patchGCN,DSMIL}. As a single WSI often contains billions of pixels and patch-level annotations are labour-intensive~\cite{lu2021clam}, a common practice is to tile the WSI into patches and utilize multi-instance learning (MIL) for classification.
Multi-instance learning is
a weakly-supervised paradigm in which each slide (bag) comprises many patches (instances)~\cite{DSMIL,qu2022weno}. 
\begin{figure}
    \centering
    \includegraphics[width=0.9\linewidth]{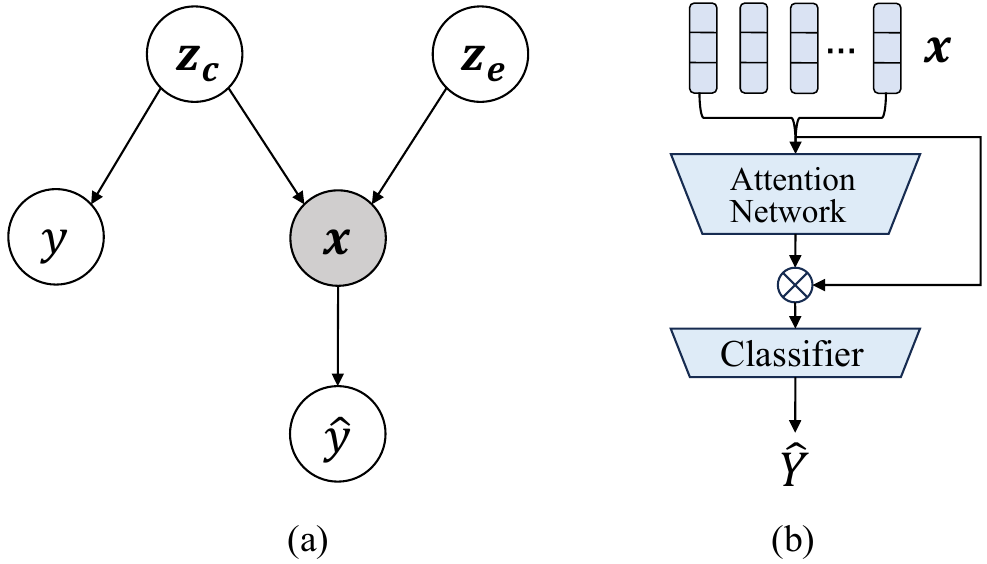}
    \caption{(a) A causal graph illustrating the generative procedure of instances and their labels. Ideally, $\bm{z}_c$ is used to predict $y$. (b) The architecture of attention-based MIL (excluding the bag feature for brevity), where the classifier can leverage any of $\bm{z}_c$ and/or $\bm{z}_e$ for prediction.}
    \label{figure:1}
\end{figure}

Most existing MIL-based WSI classification methods focus on slide-level predictions, using attention mechanisms~\cite{vaswani2017attention} to aggregate patch-level representations~\cite{ilse2018ABMIL,DSMIL,shao2021transmil,qu2022weno,zhang2022dtfd,zhang2023attention,tang2023demipl,tang2023multiple,tang2024elimipl,tang2024R2T}.
As illustrated in Figure~\ref{figure:1} (b), they assign attention weights to instances within a bag and aggregate weighted instance representations to a bag representation for classification. 
However, recent studies have revealed that these attention-based methods are prone to relying on spurious correlations as shortcuts~\cite{lin2023IBML,kaczmarzyk2024explainable,kheiri2025investigation}, attending to various biases introduced during digital slide preparation processes. This harms out-of-distribution (OOD) generalization~\cite{lin2023IBML,castro2020causality,li2020accurate} and produces misleading localization by assigning high attention to non-tumor regions while missing small tumor ones~\cite{lin2023IBML,javed2022additive,kaczmarzyk2024explainable}.

Figure~\ref{figure:1}~(a) illustrates a plausible causal structure underlying the generation process of a patch $\bm{x}$, its ground-truth label $y$, and its predicted label $\hat{y}$. We posit that patch features in WSIs are determined by two sets of factors: content factor $\bm{z}_c$, which represent the intrinsic morphological information of various tissues and cells, and environmental or bias-related factors $\bm{z}_e$, which encompass various factors introduced during slide preparation processes, including hematoxylin and eosin staining conditions, scanning equipment variations~\cite{duenweg2023whole}, etc. 
For $\bm{z}_c$, we further decompose it into \textit{causal factors $\bm{z}_{causal}$} that determine positive labels (e.g., cancerous cell morphology) and \textit{negative content factors} $\bm{z}_{neg}$ that determine negative labels (e.g., normal cell morphology).

The fundamental unreliability of attention-based MIL stems from its holistic aggregation strategy. As shown in Figure~\ref{figure:1}~(b), the bag-level representation used for classification in attention models is formed by weighted aggregation of all instances, allowing the model to freely learn correlations between various patterns and positive or negative labels to fit the training data~\cite{raff2023reproducibilityinMIL}. The behavior of learning positive and negative features is similar to standard supervised learning. However, the standard MIL framework exhibits an inherent asymmetry: \textit{only the positive concept serves as the cause of the bag label}. Specifically, its presence leads to a positive label, while its 
absence leads to a negative label. The presence of negative concepts is irrelevant to the bag label~\cite{foulds2010review,dietterich1997solving}.

This paper aims to propose a \textbf{theoretically grounded} MIL method. We turn our attention to max-pooling-based methods, which align more closely with the standard MIL setting. We provide a novel understanding of max-pooling methods from the perspective of causal invariance. Our theoretical analysis in Section \ref{sec:theory} demonstrates that, under the assumption of environmental co-occurrence in both positive and negative bags, max-pooling-based models cannot rely on environmental factors for prediction. Under the assumption of negative content co-occurrence, the model focuses solely on environment-invariant causal factors $\bm{z}_{causal}$ for prediction, ignoring environment factor $\bm{z}_e$  and negative content factors $\bm{z}_{neg}$. We also provide a thorough discussion of cases where these assumptions are violated.

A natural question arises: why do existing max-pooling-based methods underperform on WSI tasks~\cite{DSMIL,shi2020loss,shao2021transmil,zhang2022dtfd}? Through empirical investigation in Section~\ref{max-underperform}, we discover that max-pooling-based methods tend to fit the training set by memorizing instance features within the bag, resulting in complete failure in the validation set. 

We leverage variational information bottleneck (VIB)~\cite{alemi2017vib} to address the problem of naive memorization. By applying VIB to constrain the complexity of instance-level latent representation $\boldsymbol{z}$ while remaining maximally expressive about the label, we introduce an information bottleneck ~\cite{tishby2015informationbottleneck} that prevents neural networks from memorizing the instance feature flexibly. Another benefit of introducing VIB is filtering out minor noise and idiosyncratic perturbations in instances, further enhancing model robustness against adversarial inputs~\cite{alemi2017vib}. We name our method FocusMIL, reflecting two complementary perspectives: (1) VIB encourages the model to \textbf{focus} on learning concise and noise-resistant representations;  
(2) Max-pooling constrains the model to \textbf{focus} on causal factors $\bm{z}_{causal}$ while ignoring non-causal factors $\bm{z}_e$ and $\bm{z}_{neg}$.    \looseness=-1

Since max-pooling involves only one instance in a training iteration for each bag, the training can be unstable. When encountering hard instances, such as low-grade tumour cells that are morphologically similar to normal cells, the model may still overfit by memorizing incidental features instead of learning true causal patterns. To mitigate this, we introduce a mini-batch gradient descent strategy that simultaneously optimizes multiple bags in a batch. Besides the advantage of training speed, we find that this strategy effectively improves the detection of hard instances.

We validate our method on three real-world datasets and one semi-synthetic dataset. Experimental results demonstrate that FocusMIL substantially outperforms attention-based baselines under domain shift. FocusMIL achieves significantly better results compared to attention-based models in tumour region localization and patch-level prediction tasks, and remains competitive on slide-level classification. Overall, FocusMIL offers a more robust and reliable path to clinically meaningful WSI analysis.

\section{Related Work}
\subsection{Attention-based MIL for WSI classification}

Most MIL-based WSI classification algorithms aggregate patch-level features into bag-level features according to their attention scores. An early bag-based method, ABMIL~\cite{ilse2018ABMIL}, utilizes a trainable network to calculate the attention scores of instance features for bag-level classification and obtains a bag-level representation through their weighted sum. Subsequent methods have inherited similar ideas and extensively incorporated various attention mechanisms~\cite{DSMIL,lu2021clam,shao2021transmil,zhang2022dtfd,tang2024miplma}. Some of these methods try to explore spatial contextual information. In tumour pathology, the surrounding microenvironment can help with diagnosis.

Attention-based MIL methods face the challenge of overfitting from three sources~\cite{zhang2022dtfd,lin2023IBML,tang2023multiple,zhang2023attention}. Firstly, 
many cancerous slides contain \emph{salient instances}, i.e., areas with highly differentiated cancer cells that significantly differ from the normal ones~\cite{qu2024rethinking,jogi2012cancer}. 
Secondly, other slides may have \emph{hard instances}, e.g., small cancerous areas where the cancer cells closely resemble normal cells~\cite{bejnordi2017Camelyon16}.  
Thirdly, the digital scanning equipment, slide preparation, and staining processes often introduce \emph{various sources of bias} into WSI datasets~\cite{zhang2022benchmarking,sikaroudi2023generalization}. 

Existing studies~\cite{tang2023multiple,zhang2023attention} have indicated that attention models may excessively focus on salient instances and fail to identify hard instances. For positive slides containing only hard instances, these methods often rely on spurious correlations caused by bias-related factors to fit the training set~\cite{lin2023IBML}. The reliance on bias-related factors severely impacts the generalization performance of these methods, hindering the application of automatic WSI analysis in real-world scenarios.

Some recent studies have attempted to propose plug-and-play frameworks to address problems faced by attention-based MIL heads. WENO~\cite{qu2022weno} introduces knowledge distillation into WSI classification. MHIM-MIL~\cite{tang2023multiple} enhances recognition of hard instances by masking the salient ones. IBMIL~\cite{lin2023IBML}, CaMIL~\cite{chen2024camil} and CATTMIL~\cite{wu2024causal} integrate debiasing techniques from causal inference to strengthen resistance against spurious correlations. ACMIL~\cite{zhang2023attention} and DTFD-MIL~\cite{zhang2022dtfd} aim to mitigate the problem of attention overly concentrating on salient instances. Li \textit{et al.}~\cite{li2023task} introduced a three-stage task-specific fine-tuning framework using the VIB for WSI classification. In their experimental results, after fine-tuning the backbone network, the max-pooling method demonstrates superior performance to attention methods; however, they do not conduct an in-depth discussion of this phenomenon.

Regarding the reliability issues with existing MIL heads, our work proposes a reliable MIL head that directly predicts at the instance level and remains robust to domain shifts.

\subsection{Reproducibility and Learnability in Multiple Instance Learning}
Recent studies~\cite{raff2023reproducibilityinMIL} have shown that many widely used MIL methods, such as MI-Net~\cite{wang2018revisiting} and TransMIL~\cite{shao2021transmil}, do not adhere to the standard MIL assumptions and may leverage inductive biases, such as the absence of certain instances, as signals for predicting positive bags.
As various sources of bias, including staining variability and preparation artifacts, are prevalent in WSI analysis~\cite{lin2023IBML,zhang2022benchmarking}, models that do not respect the standard MIL assumptions can learn to exploit these spurious correlations~\cite{raff2023reproducibilityinMIL}. Kaczmarzyk \textit{et al.} ~\cite{kaczmarzyk2024explainable} found that attention-based methods heavily rely on non-tumour regions for metastasis detection and often neglect small tumour regions. We provide a comprehensive discussion in the Appendix on why attention-based MIL methods tend to rely on spurious correlations.

mi-Net~\cite{zhou2002neural,wang2018revisiting}, and the recently proposed CausalMIL~\cite{zhang2022CausalMIL} are perhaps the only two methods that respect the MIL assumptions. CausalMIL applies a variational autoencoder~\cite{kingma2013VAE} to the MIL problem, using a non-factorized prior distribution conditioned on bag information to provide a theoretical guarantee for identifying latent causal representation.
However, mi-Net does not perform optimally in WSI classification, and CausalMIL has not yet been studied for WSI classification.

Furthermore, the latest research on the learnability~\cite{jang2024MIL_learnable} of MIL has theoretically analyzed that all attention-based methods are unlearnable at the instance-level, whereas instance-based approaches are learnable. This further highlights the need for cautious use of attention-based MIL methods and advocates for the exploration of max-pooling-based approaches.

\section{Revisiting Max-Pooling-Based MIL: A Causal Perspective} \label{sec:theory}

\subsection{Multi-Instance Learning (MIL)}

A WSI dataset with $n$ slides are treated as MIL bags $\mathcal{B} = \{\mathbf{B}_1, \cdots, \mathbf{B}_i, \cdots, \mathbf{B}_n\}$, where each slide has a slide-level label $Y_i \in \{0, 1\}$ during training. Each slide is then cropped into patches corresponding to MIL instances $\mathbf{B}_i = \{\boldsymbol{p}_{i1}, \cdots, \boldsymbol{p}_{ij}, \cdots, \boldsymbol{p}_{in_i}\}$, where $n_i$ is the number of patches in the slide. For each patch $\boldsymbol{p}_{ij}$, there exists a patch-level label $y_{ij} \in \{0, 1\}$ that is unknown to the learner. The standard multi-instance assumption states that a bag is positive if and only if at least one of its instances is positive:
\begin{equation}
Y_i=\begin{cases}
0, & \text{if } \sum_j y_{ij}=0, \\
1, & \text{otherwise},
\end{cases}
\end{equation}
which is equivalent to using max-pooling on the instances within a bag:
\begin{equation}
Y_i = \max_j \{ y_{ij} \}.
\end{equation}
Given a feature extractor $h$, each instance is projected onto a $d$-dimensional feature vector $\boldsymbol{x}_{ij} = h(\boldsymbol{p}_{ij}) \in \mathbb{R}^d$.



\subsection{A Causal Perspective on Max-Pooling MIL}


\subsubsection{Formal Framework and Assumptions}


Following the DAG illustrated in Figure~\ref{figure:1}(a),
we formalize the data-generating process underlying MIL.
Each instance feature $\boldsymbol{x}_{ij}$ 
is generated from two latent factors:

\textbf{(1) Content factor.}

\begin{equation}
\bm{z}_{c,ij} \in \{\bm{z}_{\text{causal}}, \bm{z}_{n_1}, \ldots, \bm{z}_{n_k}\} \subset \mathbb{R}^{d_c}, \quad d_c \ll d,
\label{eq:zcontent}
\end{equation}
where $\bm{z}_{\text{causal}}$ represents the \textbf{causal latent factor} 
(e.g., latent representation of tumour morphology) that causes positive labels, 
and $\bm{z}_{n_1}, \ldots, \bm{z}_{n_k}$ represent $k$ distinct \textbf{negative latent factors} 
(e.g., latent representations of various normal tissue types) that correspond to negative labels.


\textbf{(2) Environmental factor.}
\begin{equation}
\bm{z}_{e,ij} \in \mathcal{E} \subset \mathbb{R}^{d_e}, \quad d_e \ll d,
\label{eq:ze}
\end{equation}
where $\mathcal{E}$ denotes the set of all possible environmental conditions (e.g., staining variations, lighting conditions, scanner types).
The bag-level environment defined as is 
\begin{equation}
Z_{E,i} = \{\bm{z}_{e,ij}\}_{j=1}^{n_i} \subseteq \mathcal{E},
\label{eq:ZE}
\end{equation}
as all instances within a slide share the same global preparation conditions, while still allowing for local variation.

The generative process is defined as
\begin{equation}
\boldsymbol{x}_{ij} = g(\bm{z}_{c,ij}, \bm{z}_{e,ij}, \epsilon_{ij}),
\label{eq:gen}
\end{equation}
where $g$ is an unknown generative function, and $\epsilon_{ij}$ is random noise independent of $\bm{z}_c$ and $\bm{z}_e$.

Given the generative process, the max-pooling-based model uses a feed-forward neural network \(f_\theta\) to map each instance into instance scores $s_\theta$, which are then pooled to form the bag-level score $S_\theta$:
\begin{equation}
\begin{aligned}
s_\theta(\boldsymbol{x}_{ij}) &= \big(f_\theta\!\circ g\big)\!\left(\boldsymbol{z}_{c,ij}, \boldsymbol{z}_{e,ij}, \epsilon_{ij}\right),\\
S_\theta(\mathbf{B}_i) &= \max_{j} s_\theta(\boldsymbol{x}_{ij}).
\end{aligned}
\label{eq:score}
\end{equation}
The model is trained by minimizing the empirical cross-entropy risk at the bag-level
\begin{equation}
\begin{aligned}
\mathcal{R}_n(\theta)
  = \frac{1}{n}\sum_{i=1}^n
   \Big[&-Y_i \log S_\theta(\mathbf{B}_i)\\
         &-(1-Y_i)\log(1-S_\theta(\mathbf{B}_i))\Big].
\end{aligned}
\label{eq:risk}
\end{equation}

To characterize the conditions under which bag-level supervision 
enables the model to disentangle causal content from environmental 
variation, we introduce a set of assumptions that guarantee 
identifiability and support causal generalization.


\noindent\textbf{A1 (Standard MIL Assumption).}  
A bag is positive iff it contains at least one positive instance:
\begin{equation}
Y_i = 1 \Leftrightarrow \exists \boldsymbol{x}_{ij}\in \mathbf{B}_i \text{ s.t. } y_{ij}=1.
\label{eq:mil}
\end{equation}
\noindent\textbf{A2 (Independent Causal Mechanisms).}  
Whether an instance is positive depends only on its content factor $\bm{z}_c$:
\begin{equation}
P(y \mid \bm{z}_c, \bm{z}_e) = P(y \mid \bm{z}_c),
\label{eq:icm}
\end{equation}
which reflects the principle of Independent Causal Mechanisms (ICM)~\cite{peters2017elements}: causal features determine the label, while nuisance factors vary independently.

\noindent\textbf{A3 (Separability).}
There exists a parameter setting that separates positive and negative instances with margin:
\begin{equation}
\begin{aligned}
\exists\,\tilde{\bm{\theta}},\;\alpha>\beta\in(0,1)\!:\;
s_{\tilde{\bm{\theta}}}(\boldsymbol{x})\!\ge\!\alpha\;&\text{ if }y=1,\\
s_{\tilde{\bm{\theta}}}(\boldsymbol{x})\!\le\!\beta\;&\text{ if }y=0.
\end{aligned}
\label{eq:sep}
\end{equation}

\noindent\textbf{A4 (Multi-Environment Co-occurrence).}  
(1) Positive bags come from at least two different environments:  
$\exists\,\mathbf{B}_i,\mathbf{B}_j\!\in\!\mathcal{B}^+\text{ s.t. }Z_{E,i}\!\ne\!Z_{E,j}$;  
(2) every environments seen in positive bags must also appear in negative ones:
\begin{equation}
\bigcup_{\mathbf{B}_i\in\mathcal{B}^+}\!Z_{E,i}
   \subseteq
 \bigcup_{\mathbf{B}_j\in\mathcal{B}^-}\!Z_{E,j}.
\label{eq:multi_env}
\end{equation}

\noindent\textbf{A5 (Negative Concept Co-occurrence).}  
Non-causal content factor (e.g., normal tissue morphology) 
that appears in positive bags must also appear in negative bags.  
Let $\mathcal{C}_{\text{neg}}^+$ and 
$\mathcal{C}_{\text{neg}}^-$ 
denote the sets of negative latent content factors occurring in positive and negative bags, respectively:
\begin{equation}
\begin{aligned}
\mathcal{C}_{\text{neg}}^+ &= \bigcup_{\mathbf{B}_i \in \mathcal{B}^+} \{\bm{z}_{c,ij} \mid \boldsymbol{x}_{ij} \in \mathbf{B}_i, \bm{z}_{c,ij} \in \{\bm{z}_{n_1}, \ldots, \bm{z}_{n_k}\}\},\\
\mathcal{C}_{\text{neg}}^- &= \bigcup_{\mathbf{B}_j \in \mathcal{B}^-} \{\bm{z}_{c,jk} \mid \boldsymbol{x}_{jk} \in \mathbf{B}_j, \bm{z}_{c,jk} \in \{\bm{z}_{n_1}, \ldots, \bm{z}_{n_k}\}\}. 
\end{aligned}
\label{eq:neg_sets}
\end{equation}
We assume
\begin{equation}
\mathcal{C}_{\text{neg}}^+ \subseteq \mathcal{C}_{\text{neg}}^-.
\label{eq:neg_inclusion}
\end{equation}

\paragraph{Discussion of Assumptions.}  
In practice, A4 is usually easy to satisfy since slides from different preparation centers almost always contains both positive and negative samples.  
For A5, various normal tissue cell types commonly appear 
in both positive and negative slides~\cite{bejnordi2017Camelyon16,litjens20181399}.  
We discuss cases where A5 may be violated in the Appendix~\ref{sec:appendix_assumptions_discussion}.

\subsubsection{Theoretical Analysis}
We now present our main theoretical results. Our analysis proceeds in 
four steps: (1) Lemmas~\ref{lem:opt} and~\ref{lem:locality} characterize the gradient behavior of max-pooling, showing its directional 
properties and sparsity. 
(2)
Lemma~\ref{lem:unlearnable} establishes that the model cannot treat negative content factors or environmental factors in negative bags as evidence for the negative label. 
(3) Using
Assumption A4, Theorem~\ref{thm:weak} proves that models trained via Empirical Risk Minimization (ERM) cannot exploit environmental shortcuts, yielding environmental-invariant classifiers.
(4) With Assumption A5, Theorem~\ref{thm:strong} shows 
that the model is forced to focus solely on the causal content factor. 
Together, these results provide both mechanistic insights 
and formal guarantees for max-pooling's robustness.
Due to page limit, we include proof sketches of key lemmas and theorems; full proofs are included in the supplementary material.
\begin{lemma}[Optimization Direction]\label{lem:opt}
During a gradient update step, 
the score of the argmax instance in each bag moves in the direction consistent with the bag label:
\begin{itemize}
\item For a positive bag, its argmax instance score increases.
\item For a negative bag, its argmax instance score decreases.
\end{itemize}
\end{lemma}

\begin{lemma}[Gradient Locality]\label{lem:locality}
At any parameter setting $\bm{\theta}$, non-argmax instances has zero gradient:
\begin{equation}
\frac{\partial\mathcal{L}_i}{\partial s_{\bm{\theta}}(\boldsymbol{x}_{ik})}=0,
\quad
\boldsymbol{x}_{ik}\ne \boldsymbol{x}_i^*=\arg\max_j s_{\bm{\theta}}(\boldsymbol{x}_{ij}).
\label{eq:grad_local}
\end{equation}
Hence, only the argmax instance contributes to back-propagation. 
\end{lemma}

\begin{lemma}[Unlearnability of Negative Concepts]\label{lem:unlearnable}
At any empirical risk minimum, the model cannot use negative content factors 
($\bm{z}_c \in \{\bm{z}_{n_1}, \ldots, \bm{z}_{n_k}\}$) 
or environment factor ($\bm{z}_e \in \mathcal{E}$) as signals for negative bag labels.
\end{lemma}
\noindent\textbf{Proof sketch.}  
If a negative instance with $\bm{z}_c \in \{\bm{z}_{n_1}, \ldots, \bm{z}_{n_k}\}$ 
or $\bm{z}_e = e'$ receives a high score:
\begin{itemize}
\item If it is not the argmax, it receives no gradient (Lemma~\ref{lem:locality});
\item If it is the argmax, gradient descent (Lemma~\ref{lem:opt}) decreases its score until another instance becomes dominant.
\end{itemize}
Thus, any dependence on such patterns is unstable and disappears during training.

Lemma~\ref{lem:unlearnable} highlights a key difference of MIL compared with standard supervised learning: max-pooling prevents MIL models from learning features of negative bags and forces them to focus solely on positive evidence. This reflects the inherent asymmetry of MIL, where bag labels are determined by the presence of positive instances rather than any defining negative pattern. A formal definition and complete proofs are provided in Appendix~\ref{appendix_lemma3}.


\begin{theorem}[Environmental Robustness]\label{thm:weak}
Under A1--A4, any global minimizer $\bm{\theta^*}$ of the empirical risk 
$\mathcal{R}_n(\bm{\theta})$ for a max-pooling-based model is environment-invariant, its scoring function depends only on $\bm{z}_c$.
\end{theorem}

\noindent\textbf{Proof sketch.}  
By Lemma~\ref{lem:unlearnable}, the classifier cannot use negative content factors 
($\bm{z}_c \in \{\bm{z}_{n_1}, \ldots, \bm{z}_{n_k}\}$) 
or environment factors ($\bm{z}_e \in \mathcal{E}$) as signals for negative labels. 
For positive bags, if the classifier relies on an environment $\bm{z}_e = e'$ to assign high scores, 
A4 guarantees that $e'$ also appears in negative bags, forcing high scores and increasing the loss, 
contradicting ERM optimality. 


\begin{theorem}[Focusing on Causal Factors]\label{thm:strong}
Under A1--A5,  any global minimizer $\bm{\theta}^*$ of the empirical risk 
$\mathcal{R}_n(\theta)$ for a max-pooling-based model yields an invariant \emph{causal classifier}.
Its instance-level scoring function $s_{\theta^*}(\boldsymbol{x})$ depends solely on the causal factor
$\bm{z}_{\text{causal}}$ for prediction, while ignoring environmental factors ($\mathcal{E}$) 
and non-causal content factors ($\{\bm{z}_{n_1}, \ldots, \bm{z}_{n_k}\}$).
\end{theorem}

\noindent\textbf{Proof sketch.}  
By Theorem~\ref{thm:weak}, the classifier already ignores environmental factors.
If the classifier assigns high scores to instances containing non-causal content factors $\bm{z}_c \in \{\bm{z}_{n_1}, \ldots, \bm{z}_{n_k}\}$,
A5 ensures that these factors also appear in negative bags,
which leads to high scores and increases the loss, contradicting ERM optimality.
Therefore, the model must rely solely on the causal factor $\bm{z}_c = \bm{z}_{\text{causal}}$.


\subsection{Why Existing Max-Pooling-Based Models Underperform?}\label{max-underperform}


Despite the favorable theoretical properties discussed above, existing max-pooling MIL algorithms~\cite{wang2018revisiting} often underperform and have received relatively little attention in recent years. 
Through empirical investigation, we discover that when positive concepts are not strongly separable, these models tend to memorize instance features within each bag rather than learn meaningful positive patterns. They can fit the training set perfectly yet produce near-random validation predictions and fail almost entirely at patch-level inference (see Appendix~\ref{sec:rote-memorization} for empirical evidence). This reflects the well-known memorization behavior of neural networks, which can even fit data with random labels~\cite{wei2024memorization}. Using stronger feature extractors (e.g., CTransPath~\cite{wang2022ctranspath}) can mitigate these issues by making positive concepts more distinguishable.

The second problem is that deterministic max-pooling depends entirely on a single top-scoring instance, making the model unstable and highly sensitive to incidental perturbations or noise in individual instances. For slides with only small tumor regions (i.e., hard instances), normal patches can be mistakenly selected as the argmax, leading to erroneous gradient updates. As a result, standard max-pooling MIL tends to focus only on salient, easy instances while failing to detect subtle or hard positive patches, a weakness that is especially problematic in early-stage or mild pathological cases.

\begin{figure*}[htbp]
    \centering
    \includegraphics[width=0.85\textwidth]{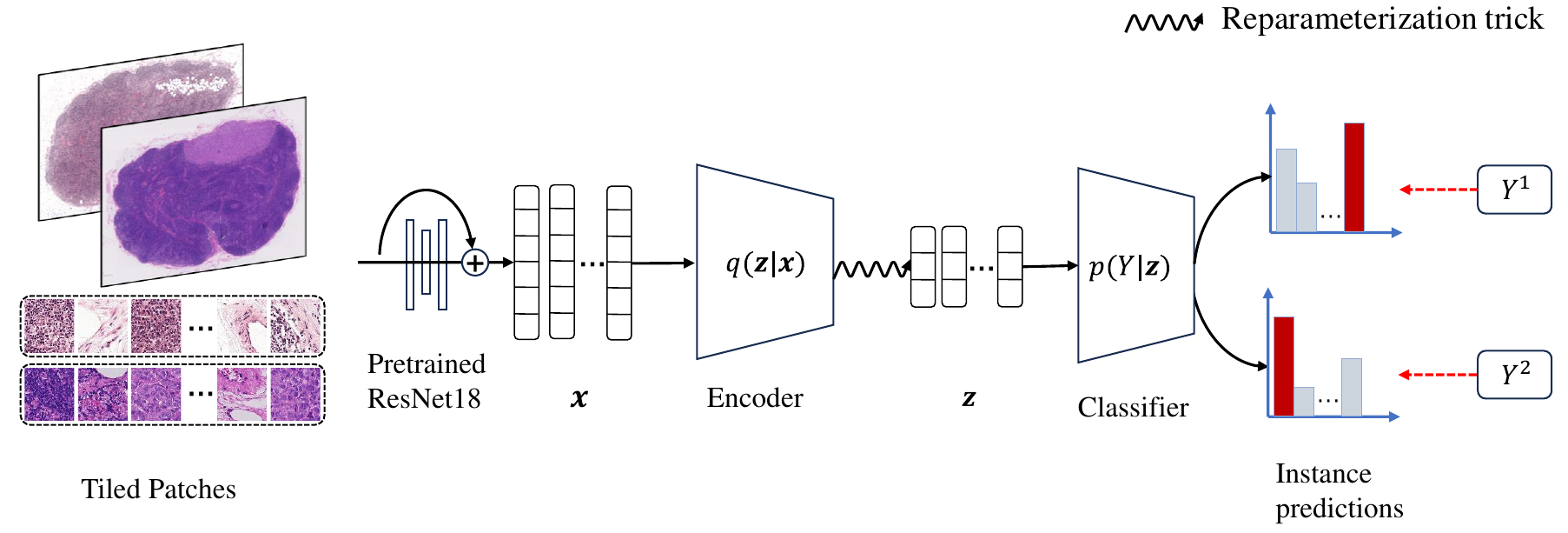}
    \caption{Overview of our FocusMIL with 2 slides in a mini-batch for illustration. Features are extracted from patches using a pre-trained feature extractor. The instance features are encoded into latent representations via a stochastic encoder. The classifier $P(Y|\bm{z})$ uses the inferred latent representations \(\boldsymbol{z} \) to obtain patch-level prediction scores. Max-pooling is applied to the instance predictions of each MIL bag to derive the slide-level prediction.}
    \label{fig:4}
\end{figure*}
\section{FocusMIL}
To address the two aforementioned challenges, we propose FocusMIL, a max-pooling-based model regularised by the VIB and trained with a multi-slide mini-batch strategy. Figure~\ref{fig:4} illustrates the architecture of our model.
\paragraph{Variational Information Bottleneck Regularization.} Following the information-bottleneck principle~\cite{tishby99information}, we aim to learn a latent representation $\boldsymbol{z}$ that is maximally predictive of the label \(Y\) while being maximally compressive of the input $\boldsymbol{x}$. We can maximize the objective function:
\begin{equation}
\mathcal{R}_{\text{IB}}(\theta)
\;=\;
I_\theta(Z;Y)\;-\;\beta\,I_\theta(Z;X),
\label{eq:ib}
\end{equation}

where $I(\cdot; \cdot)$ denotes mutual information, $\theta$ the network parameters and $\beta$ controls the trade-off between predictive power and compression. Intuitively, the first term in Equation~\ref{eq:ib} encourages $Z$ to be predictive of \(Y\) while the second term forces $Z$ to forget the complex details in the input.

Since the computation of $I_\theta(Z;X)$ is intractable, we follow the VIB framework~\cite{alemi2017vib}.
Specifically, we approximate \(I_\theta(Z;X)\) via variational inference
and train the network by minimising a tractable variational upper bound.
For simplicity, we set the prior distribution as $p(\bm{z})=\mathcal{N}(0,\bm{I})$. 
The final loss function is:
\begin{equation}
\mathcal{L} = -\log p\left(Y \mid \boldsymbol{z}^*\right) + \beta \cdot \operatorname{KL}\left[q\left(\boldsymbol{z} \mid \boldsymbol{x}\right) \| p\left(\boldsymbol{z}\right)\right]
\end{equation}
where $\beta$ is a hyperparameter for balancing the classifier and KL loss, $\boldsymbol{z}^*$ corresponds to the most positive latent representation of a bag obtained with max-pooling over $p(Y|\bm{z})$. A large $\beta$ enforces stronger regularization, which may
degrade predictive performance, whereas a small $\beta$ may fail to prevent the network from naive memorization.

An additional benefit of introducing the VIB regularization is robustness to idiosyncratic perturbations: because the encoder outputs a distribution
rather than a deterministic code, slide-specific noise and artefacts are
unlikely to pass through the latent bottleneck. This
property is crucial for WSI classification, where training slides are often scarce and max-pooling relies on just one patch from each slide during optimization, making conventional networks prone to overfitting on perturbations.

The key difference between our method and CausalMIL~\cite{zhang2022CausalMIL} is that our approach regularizes the latent space of all instances, whereas CausalMIL only regularizes the latent space of the most positive instance per bag. Our subsequent experiments demonstrate that CausalMIL cannot effectively mitigate the simple memorization of bags on more complex datasets. Additionally, since we believe that max-pooling operation inherently enforces the disentangling of $\bm{z}_c$ and $\bm{z}_e$, our method does not require instance reconstruction.

\paragraph{Multi-Slide Mini-Batch Gradient Descent for WSI Classification.}

To address the training instability discussed in Section~\ref{max-underperform}, we place multiple slides in a single mini-batch, performing forward propagation and optimization simultaneously.

This strategy provides two benefits: (1) Gradient updates are averaged across all bags. Salient tumor instances within the batch generate gradients toward genuine tumor features, which mitigate the misleading gradients arising from mistakenly selected instances in hard samples. Moreover, negative bags provide an opposing force: when similar normal patterns appear as argmax in negative bags, their gradients push scores downward, counteracting the erroneous upward push from hard positive bags. (2) Once salient tumor features have been learned, the model is further encouraged to explore shared tumor representations that can simultaneously recognize both salient and hard instances across different slides, thereby establishing more generalizable classification boundaries.

We empirically found that mini-batch optimization leads to better recognition of hard instances, as confirmed by improved tumor region localization (FROC) and visualization results in our experiments.


\begin{table*}[htbp]
\centering
\caption{Performance of MIL methods using ResNet-18 features on Camelyon16 Dataset.
The subscripts show the \(95\%\) confidence intervals.
The best result is in \textbf{bold}; the second-best is \underline{underlined}.}
\label{tab:camelyon16}
\resizebox{0.82\textwidth}{!}{
\begin{tabular}{lcccc}
\toprule
\multirow{2}{*}{Method}
& \multicolumn{2}{c}{Slide-level} &
  \multicolumn{2}{c}{Patch-level} \\ \cmidrule(r){2-3} \cmidrule(r){4-5}
& AUC & ACC & AUCPR & F1-score \\
\midrule
ABMIL     & 0.8052 \scriptsize{(0.7492, 0.8612)} & 0.8152 \scriptsize{(0.7743, 0.8560)} & 0.1449 \scriptsize{(0.0449, 0.2449)} & 0.1570 \scriptsize{(0.0658, 0.2482)} \\
DSMIL     & 0.7733 \scriptsize{(0.7167, 0.8300)} & 0.7984 \scriptsize{(0.7653, 0.8316)} & 0.2693 \scriptsize{(0.0409, 0.4976)} & 0.2468 \scriptsize{(0.1147, 0.3790)} \\
TransMIL  & 0.8352 \scriptsize{(0.7807, 0.8898)} & 0.8127 \scriptsize{(0.7865, 0.8390)} & -- & -- \\
DTFD-MIL  & \underline{0.8619} \scriptsize{(0.8530, 0.8709)} & 0.8032 \scriptsize{(0.7497, 0.8567)} & 0.2079 \scriptsize{(0.1592, 0.2566)} & 0.1640 \scriptsize{(0.1388, 0.1891)} \\
IBMIL     & 0.8442 \scriptsize{(0.8329, 0.8555)} & 0.7906 \scriptsize{(0.7465, 0.8347)} & 0.2587 \scriptsize{(0.2138, 0.3037)} & 0.1948 \scriptsize{(0.1499, 0.2398)} \\
CATTMIL     & 0.8231 \scriptsize{(0.8070, 0.8393)} & 0.8035 \scriptsize{(0.7898, 0.8172)} & 0.2776 \scriptsize{(0.1522, 0.4030)} & 0.2754 \scriptsize{(0.1785, 0.3723)} \\
AEM         & 0.8111 \scriptsize{(0.7999, 0.8223)} & 0.8187 \scriptsize{(0.8038, 0.8336)} & 0.2670 \scriptsize{(0.2148, 0.3192)} & 0.3264 \scriptsize{(0.2879, 0.3649)} \\
Conjunctive & 0.8119 \scriptsize{(0.8020, 0.8218)} & \underline{0.8297} \scriptsize{(0.7987, 0.8607)} & 0.3703 \scriptsize{(0.3393, 0.4013)} & \underline{0.3950} \scriptsize{(0.3441, 0.4459)} \\
mi-Net    & 0.8014 \scriptsize{(0.7674, 0.8355)} & 0.8158 \scriptsize{(0.7900, 0.8416)} & 0.4108 \scriptsize{(0.3466, 0.4751)} & 0.3844 \scriptsize{(0.3437, 0.4251)} \\
CausalMIL & 0.8092 \scriptsize{(0.7730, 0.8454)} & 0.8281 \scriptsize{(0.8128, 0.8435)} & \textbf{0.4845} \scriptsize{(0.4653, 0.5036)} & 0.3650 \scriptsize{(0.3469, 0.3823)} \\
FocusMIL  & \textbf{0.8706} \scriptsize{(0.8489, 0.8923)} & \textbf{0.8359} \scriptsize{(0.8206, 0.8513)} & \underline{0.4382} \scriptsize{(0.3907, 0.4857)} & \textbf{0.3996} \scriptsize{(0.3787, 0.4205)} \\
\bottomrule
\end{tabular}}
\end{table*}

\begin{table*}[htbp]
\centering
\caption{Performance of MIL methods on Camelyon17 and TCGA-NSCLC.}
\label{tab:cam17_tcga}
\resizebox{0.82\textwidth}{!}{
\begin{tabular}{lcccc}
\toprule
\multirow{2}{*}{Method} & \multicolumn{2}{c}{Camelyon17} & \multicolumn{2}{c}{TCGA-NSCLC} \\ 
\cmidrule(lr){2-3}\cmidrule(lr){4-5}
& AUC & F1-score & AUC & ACC \\
\midrule
ABMIL     & 0.7877 \scriptsize{(0.7505, 0.8249)} & 0.7004 \scriptsize{(0.6702, 0.7306)} & 0.8900 \scriptsize{(0.8739, 0.9061)} & 0.8200 \scriptsize{(0.8101, 0.8299)} \\
DSMIL     & 0.7069 \scriptsize{(0.5669, 0.8469)} & 0.6507 \scriptsize{(0.5826, 0.7187)} & 0.8909 \scriptsize{(0.8872, 0.8946)} & 0.8295 \scriptsize{(0.8208, 0.8382)} \\
TransMIL  & 0.6899 \scriptsize{(0.6187, 0.7610)} & 0.6200 \scriptsize{(0.5459, 0.6941)} & \textbf{0.9501} \scriptsize{(0.9476, 0.9526)} & \textbf{0.8886} \scriptsize{(0.8811, 0.8961)} \\
DTFD-MIL  & 0.8192 \scriptsize{(0.7964, 0.8419)} & 0.7242 \scriptsize{(0.6960, 0.7524)} & 0.9340 \scriptsize{(0.9328, 0.9352)} & 0.8695 \scriptsize{(0.8670, 0.8720)} \\
IBMIL     & 0.8221 \scriptsize{(0.8059, 0.8383)} & 0.7233 \scriptsize{(0.7022, 0.7444)} & \underline{0.9401} \scriptsize{(0.9389, 0.9413)} & 0.8743 \scriptsize{(0.8693, 0.8793)} \\
AEM         & 0.8005 \scriptsize{(0.7831, 0.8179)} & 0.7021 \scriptsize{(0.6611, 0.7431)} & 0.9138 \scriptsize{(0.9038, 0.9238)} & 0.8371 \scriptsize{(0.8243, 0.8499)} \\
CATTMIL     & 0.8162 \scriptsize{(0.7628, 0.8696)} & 0.7179 \scriptsize{(0.6732, 0.7626)} & 0.9159 \scriptsize{(0.9022, 0.9296)} & 0.8600 \scriptsize{(0.8405, 0.8794)} \\
Conjunctive & \underline{0.8322} \scriptsize{(0.8223, 0.8421)} & \underline{0.7378} \scriptsize{(0.7105, 0.7651)} & 0.9188 \scriptsize{(0.9099, 0.9277)} & 0.8714 \scriptsize{(0.8655, 0.8773)} \\
mi-Net    & 0.5066 \scriptsize{(0.5022, 0.5110)} & 0.2901 \scriptsize{(0.2419, 0.3383)} & 0.9314 \scriptsize{(0.9255, 0.9374)} & 0.8829 \scriptsize{(0.8699, 0.8958)} \\
CausalMIL & 0.5104 \scriptsize{(0.4615, 0.5592)} & 0.4905 \scriptsize{(0.3881, 0.5929)} & 0.9298 \scriptsize{(0.9273, 0.9323)} & 0.8791 \scriptsize{(0.8704, 0.8878)} \\
FocusMIL  & \textbf{0.8719} \scriptsize{(0.8619, 0.8818)} & \textbf{0.8335} \scriptsize{(0.8273, 0.8398)} & 0.9350 \scriptsize{(0.9316, 0.9384)} & \underline{0.8857} \scriptsize{(0.8815, 0.8899)} \\
\bottomrule
\end{tabular}}
\end{table*}

\section{Experiments}
\subsection{Datasets and Baselines}

\textbf{Camelyon16 dataset} is widely used for metastasis detection in breast cancer~\cite{bejnordi2017Camelyon16}. The dataset consists of 270 training and 129 testing WSIs.\\
\textbf{Camelyon17 dataset} consists of 1,000 WSIs from five hospitals. Due to the absence of labels in the test set, we only used the 500 slides from the training set. Following the setup of AEM~\cite{zhang2024aem}, we designated 200 slides from the fourth and fifth hospitals as the test set to evaluate the model's OOD generalization performance. The remaining 300 slides were split into training and validation sets with an 8:2 ratio.
For a more detailed discussion of the Camelyon datasets, as well as additional OOD experiment results, please refer to the Appendix.\\
\textbf{TCGA-NSCLC} includes two subtypes of lung cancer: Lung Adenocarcinoma (LUAD) and Lung Squamous Cell Carcinoma (LUSC), with a total of 1,054 diagnostic slides.\\
\textbf{Semi-Synthetic Dataset
}
To test whether existing WSI classification methods respect the standard MIL assumption and are susceptible to spurious correlations in negative bags, we propose the Camelyon16 Standard-MIL test dataset, inspired by the ideas in~\cite{raff2023reproducibilityinMIL}. Specifically, in the training set, we introduce poison by randomly selecting $20\%$ of the patches in the normal slides and increasing the intensity of their green channel. In the test set, we randomly select $20\%$ of the patches in tumour slides to introduce poison in the same way. A MIL model cannot legally learn to use the poison signal because it occurs only in normal slides~\cite{raff2023reproducibilityinMIL}. If a model has a training AUC $>0.5$, but a test AUC $<0.5$, it relies on poison for prediction and does not respect the MIL assumption.\\
\textbf{Baselines}
We compare our method with several recently published baselines, including ABMIL~\cite{ilse2018ABMIL}, DSMIL~\cite{DSMIL}, TransMIL~\cite{shao2021transmil}, DTFD-MIL~\cite{zhang2022dtfd}, IBMIL~\cite{lin2023IBML}, AEM~\cite{zhang2024aem}, CATTMIL~\cite{wu2024causal}, Conjunctive-pooling~\cite{early2024inherently}, mi-Net~\cite{wang2018revisiting}, and CausalMIL~\cite{zhang2022CausalMIL}. The first seven methods use attention mechanisms, Conjunctive-pooling combines attention with instance-level pooling, while the last three methods are based on max-pooling. To our knowledge, CausalMIL has not yet been applied for WSI classification. All experiments use ResNet-18\cite{he2016Resnet} pretrained on ImageNet\cite{deng_lifeifei2009imagenet} as the feature extractor. 

For details on evaluation metrics, results using more advanced feature extractors, ablation studies, computational cost analysis, and additional visualization results, please refer to the supplementary material.

\begin{figure*}[htbp]
    \centering
    \includegraphics[width=0.65\linewidth]{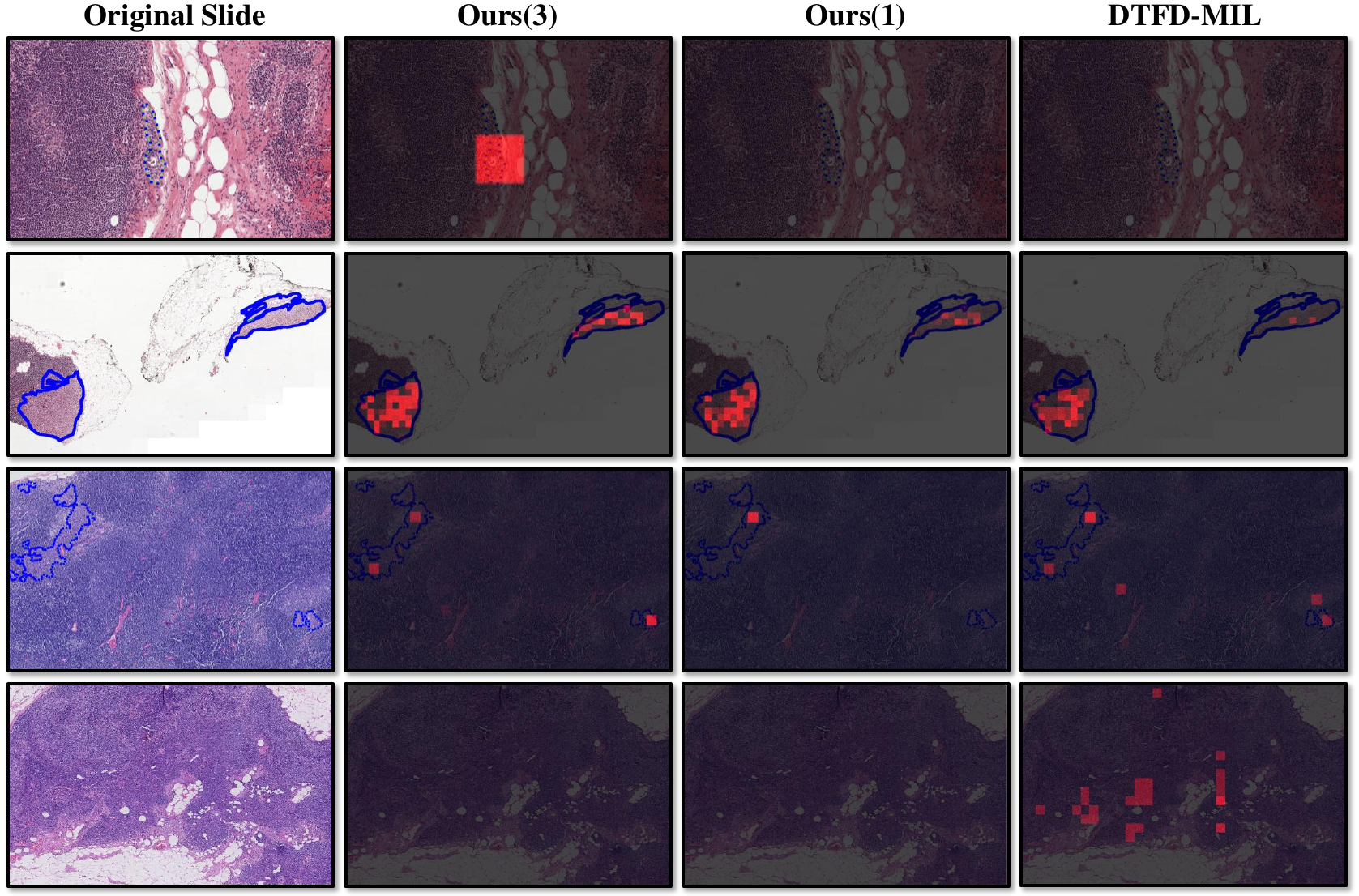}
    \caption{Visualization of FocusMIL and DTFD-MIL on Camelyon16 dataset. Ours(3) and Ours(1) refer to the FocusMIL models trained with a batch size of 3 and 1. The fourth slide is negative.}
    \label{fig:visual1}
\end{figure*}

\subsection{Results on Real-World Dataset}
\textbf{Results on Camelyon16} As shown in Table~\ref{tab:camelyon16}, for patch-level classification, max-pooling-based methods such as mi-Net, CausalMIL, and FocusMIL achieve significantly better results than the attention-based methods. For slide-level classification, mi-Net and CausalMIL achieve performance comparable to ABMIL. Our proposed FocusMIL achieves better results than the more sophisticated attention methods in terms of both AUC and ACC. For tumour region localization (Table~\ref{tab:2}), FocusMIL outperforms DTFD-MIL substantially. It is worth mentioning that after using the multi-slide mini-batch gradient optimization strategy, FocusMIL achieves better small-area tumour (hard instance) prediction compared to training with SGD. As shown in Figure~\ref{fig:visual1}, our
model accurately identifies both large and small tumour regions without any false positive predictions. FocusMIL with
a training batch size of 1 misses some small tumour regions.
For DTFD-MIL, it misses the tumour region in the first slide
and some tumour patches in the second slide, and there are
many false positive predictions.

\begin{table}[htbp]
    \centering
    \caption{Localization FROC Performance Comparison. }
    
    \begin{tabular}{l|c}
        \hline
         Method & Localization FROC \\
        \hline
        ABMIL & 0.2434 \\
        DTFD-MIL & 0.3013 \\
        FocusMIL (1) & 0.3462 \\
        FocusMIL (3) & 0.3646 \\
        \hline
    \end{tabular}
    \label{tab:2}
\end{table}

\textbf{Results on Camelyon17} Since the training and testing data come from different hospitals, the distribution of the environment information $\bm{z}_e$ may vary. In this OOD scenario,  as shown in Table~\ref{tab:cam17_tcga}, FocusMIL achieves significantly better results. However, mi-Net and CausalMIL overfit the dataset by memorizing instance features within the bag.

\textbf{Results on TCGA-NSCLC} FocusMIL achieved competitive results compared to other comparison methods. TransMIL obtained the best results by exploring spatial relationships; however, since it cannot guarantee predictions based on causal factors, it may not be reliable in actual deployment. Our method is more reliable when facing domain shifts.

\begin{table}[htbp]
\centering
\caption{Performance on Camelyon16 Standard-MIL test dataset.}
\label{tab:4}
\resizebox{0.88\columnwidth}{!}{%
\begin{tabular}{lcc}
\toprule
\textbf{Model} & \textbf{Slide AUC} & \textbf{Patch F1} \\
\midrule
ABMIL        & 0.007 \scriptsize{(0.000, 0.021)} & 0.142 \scriptsize{(0.117, 0.167)} \\
DSMIL        & 0.377 \scriptsize{(0.000, 0.776)} & 0.312 \scriptsize{(0.165, 0.459)} \\
TransMIL     & 0.001 \scriptsize{(0.001, 0.002)} & -- \\
DTFD\mbox{-}MIL & 0.010 \scriptsize{(0.000, 0.034)} & 0.122 \scriptsize{(0.080, 0.165)} \\
IBMIL        & 0.000 \scriptsize{(0.000, 0.001)} & 0.104 \scriptsize{(0.098, 0.110)} \\
CATTMIL      & 0.354 \scriptsize{(0.124, 0.584)} & 0.335 \scriptsize{(0.210, 0.460)} \\
AEM          & 0.000 \scriptsize{(0.000, 0.000)} & 0.118 \scriptsize{(0.094, 0.142)} \\
Conjunctive  & 0.000 \scriptsize{(0.000, 0.000)} & 0.099 \scriptsize{(0.097, 0.102)} \\
mi\mbox{-}Net  & 0.805 \scriptsize{(0.772, 0.838)} & \underline{0.384 \scriptsize{(0.317, 0.453)}} \\
CausalMIL    & \underline{0.808 \scriptsize{(0.785, 0.830)}} & 0.366 \scriptsize{(0.325, 0.407)} \\
FocusMIL     & \textbf{0.868 \scriptsize{(0.853, 0.883)}} & \textbf{0.384 \scriptsize{(0.366, 0.401)}} \\
\bottomrule
\end{tabular}
}
\end{table}

\subsection{Results on Standard-MIL Test}
All methods achieved a perfect slide AUC of 1.00 on the training set. On the test set, as shown in Table~\ref{tab:4},
all of the tested attention-based methods have a test slide AUC less than 0.5, failing to respect the standard MIL assumption. The three max-pooling-based methods are little affected by the poison. Benefiting from the max-pooling-based instance classifier, DSMIL performs well in patch-level classification.
Taken together, these results provide strong empirical support for Lemma~\ref{lem:unlearnable}.

\section{Conclusion}

In this paper, we revisited max-pooling-based MIL methods for WSI classification from a causal perspective. In the standard MIL framework, we reformulate the objective as learning 
positive content concepts (causal factors) which are the direct cause of bag 
labels. Under mild assumptions, we showed that max-pooling models can focus on causal content factors while ignoring environmental factors and negative content factors. We further observed that previous max-pooling-based models can fit the training data through rote memorization, leading to poor performance at test time. Motivated by these findings, we proposed \emph{FocusMIL}, which incorporates a VIB to regularize the latent space and prevent such memorization. Our experiments demonstrate that FocusMIL substantially outperforms attention-based MIL methods in OOD settings and achieves more accurate patch-level localization of tumor regions. As domain shift is common in clinical practice, we believe our work takes a substantial step toward the reliable deployment of computational pathology models.

\textbf{Future Work.}
Some WSI tasks, such as survival prediction and treatment response prediction,
may benefit from using contextual information across multiple patches within
a slide. A promising direction for future research is to develop hybrid
architectures that combine max-pooling's robustness to spurious correlations 
with attention's contextual modeling capabilities.

{
    \small
    \bibliographystyle{ieeenat_fullname}
    \bibliography{main}
}


\appendix  
\section{Theory}

\subsection{Lemma 1: Optimization Direction}

\begin{lemma}[Optimization Direction]
\label{lem:opt_direction}
During one gradient descent update step, the argmax instance in each bag changes its score consistently with the bag label:
\begin{itemize}
    \item For a positive bag ($Y_i=1$), its argmax instance score increases
    \item For a negative bag ($Y_i=0$), its argmax instance score decreases
\end{itemize}
\end{lemma}

\begin{proof}
Let $\boldsymbol{x}^*_i = \arg\max_j s_\theta(\boldsymbol{x}_{ij})$ be the argmax instance of bag $B_i$. 

Because a neural network produces real-valued scores in floating-point arithmetic, exact ties occur only when two outputs take the same floating-point value. Under typical continuous parameterizations, this has probability zero, so the argmax index is almost always unique

For the loss function:
\begin{equation}
\mathcal{L}_i(\theta) = -Y_i \log S_\theta(B_i) - (1-Y_i)\log(1-S_\theta(B_i)).
\end{equation}

Since $S_\theta(B_i) = s_\theta(\boldsymbol{x}^*_i)$, for positive bag ($Y_i=1$):
\begin{equation}
\frac{\partial \mathcal{L}_i}{\partial s_\theta(\boldsymbol{x}^*_i)} = -\frac{1}{S_\theta(B_i)} < 0,
\end{equation}
gradient descent increases the score.

For negative bag ($Y_i=0$):
\begin{equation}
\frac{\partial \mathcal{L}_i}{\partial s_\theta(\boldsymbol{x}^*_i)} = \frac{1}{1-S_\theta(B_i)} > 0,
\end{equation}
gradient descent decreases the score.
\end{proof}

\subsection{Lemma 2: Gradient Locality}

\begin{lemma}[Gradient Locality]\label{lem:locality}
At any parameter setting $\theta$, non-argmax instances have zero gradient:
\begin{equation}
\frac{\partial\mathcal{L}_i}{\partial s_\theta(\boldsymbol{x}_{ik})}=0,
\quad
\boldsymbol{x}_{ik}\ne \boldsymbol{x}_i^*=\arg\max_j s_\theta(\boldsymbol{x}_{ij}).
\label{eq:grad_local}
\end{equation}
Hence, only the argmax instance contributes to back-propagation. 
\end{lemma}

\begin{proof}
For a non-argmax instance $\boldsymbol{x}_{ik} \neq \boldsymbol{x}^*_i$, we have $s_\theta(\boldsymbol{x}_{ik}) < \max_j s_\theta(\boldsymbol{x}_{ij})$.

The partial derivative of the max function at non-maximum points is zero:
\begin{equation}
\frac{\partial S_\theta(B_i)}{\partial s_\theta(\boldsymbol{x}_{ik})} = \frac{\partial}{\partial s_\theta(\boldsymbol{x}_{ik})} \max_j s_\theta(\boldsymbol{x}_{ij}) = 0.
\end{equation}

Therefore:
\begin{equation}
\frac{\partial \mathcal{L}_i}{\partial s_\theta(\boldsymbol{x}_{ik})} = \frac{\partial \mathcal{L}_i}{\partial S_\theta(B_i)} \cdot \frac{\partial S_\theta(B_i)}{\partial s_\theta(\boldsymbol{x}_{ik})} = 0.
\end{equation}

Only the argmax instance contributes to backpropagation.
\end{proof}

\subsection{Lemma 3: Unlearnability of Negative Concepts}
\label{appendix_lemma3}

\subsubsection{Definition: Non-Causal Factor as Negative Label Signal}

Before delving into Lemma 3, we first formalize a key concept about using $\boldsymbol{z}_{\text{neg}}$ or $\boldsymbol{z}_{e}$ as signals for negative bag labels.
 
\begin{definition}[Non-Causal Factor as Negative Label Signal]
\label{def:negative_signal}
Consider a non-causal content factor $\boldsymbol{z}_{c} \in \{\boldsymbol{z}_{n_1}, \ldots, \boldsymbol{z}_{n_k}\}$ or an environment factor $\boldsymbol{e} \in \mathcal{E}$.

For any bag containing instances with this (non-causal content) factor, we can decompose the bag into two sub-bags:
\begin{itemize}
    \item $B^{(\text{base})}$: instances without this factor
    \item $B^{(\text{add})}$: instances containing this factor
\end{itemize}

We say the model ``treats this factor as a negative bag label signal'' if adding instances with this factor decreases the bag's prediction score:
\begin{equation}
S_\theta(B^{(\text{base})} \cup B^{(\text{add})}) < S_\theta(B^{(\text{base})}).
\end{equation}
\end{definition}

\paragraph{Discussion.} This means the model treats the presence of this factor as ``negative evidence'' --- when the model ``sees'' this factor, it lowers the prediction score for that bag, thus being more inclined to classify it as negative. Attention-based models can learn such shortcuts by assigning higher attention weights to instances containing bias factors, amplifying their contribution in the final bag representation, which the classifier then exploits as evidence for negative classification. For detailed analysis, see Section~\ref{sec:attention}.

\begin{lemma}[Unlearnability of Negative Concepts]
\label{lem:unlearnability}
At any empirical risk minimum, the max-pooling based model cannot use negative content factors 
($\boldsymbol{z}_c \in \{\boldsymbol{z}_{n_1}, \ldots, \boldsymbol{z}_{n_k}\}$) 
or environment factor ($\boldsymbol{z}_e \in \mathcal{E}$) as signals for negative bag labels (Definition~\ref{def:negative_signal}).
\end{lemma}

We develop the result from two perspectives. The first relies on the structural constraints imposed by max pooling, while the second follows the optimization-dynamics viewpoint outlined in the main text. These perspectives converge on the same underlying insight.

\subsubsection{Proof A: Structural Constraint}


\begin{proof}
Consider the setting in Definition~\ref{def:negative_signal}: there exist $B^{(\text{base})}$ (instances without this factor) and $B^{(\text{add})}$ (instances containing this factor).

By the definition of max-pooling:
\begin{align}
S_\theta(B^{(\text{base})} \cup B^{(\text{add})}) 
&= \max_{\boldsymbol{x} \in B^{(\text{base})} \cup B^{(\text{add})}} s_\theta(\boldsymbol{x}) \\
&= \max\left\{\max_{\boldsymbol{x} \in B^{(\text{base})}} s_\theta(\boldsymbol{x}),\ \max_{\boldsymbol{x} \in B^{(\text{add})}} s_\theta(\boldsymbol{x})\right\} \\
&= \max\{S_\theta(B^{(\text{base})}),\ S_\theta(B^{(\text{add})})\}.
\end{align}

Thus, we can get:
\begin{equation}
S_\theta(B^{(\text{base})} \cup B^{(\text{add})}) \geq S_\theta(B^{(\text{base})}).
\end{equation}

This shows the monotonicity of max-pooling: adding instances to any bag
(regardless of which factors they contain) can only increase or maintain
the bag's prediction score, never decrease it. Therefore the condition
\(
S_\theta(B^{(\text{base})} \cup B^{(\text{add})}) <
S_\theta(B^{(\text{base})})
\)
in Definition~\ref{def:negative_signal} can never hold, and no
factor can be used as a negative bag-label signal.
\end{proof}

\subsubsection{Proof B: Optimization Constraint under ERM Framework}


\begin{proof}
Assume a model with parameter $\theta$ tends to learn a certain factor $\boldsymbol{z}_c \in \{\boldsymbol{z}_{n_1}, \ldots, \boldsymbol{z}_{n_k}\}$ or $\boldsymbol{z}_e \in \mathcal{E}$ as a signal for negative labels. We analyze the two complementary cases for an instance $\boldsymbol{x}^*$ containing this factor in a negative bag during the optimization process.

\textbf{Case 1: $\boldsymbol{x}^*$ is not argmax}:

By Lemma~\ref{lem:locality}, this instance does not affect training or prediction. Therefore, the model cannot establish an association between this factor and negative labels from this instance.

\textbf{Case 2: $\boldsymbol{x}^*$ is argmax}:

In this case, the bag score is determined by $\boldsymbol{x}^*$:
\begin{equation}
S_\theta(B^-) = s_\theta(\boldsymbol{x}^*)
\end{equation}

By Lemma~\ref{lem:opt_direction}, the ERM optimization process will decrease the score of $\boldsymbol{x}^*$ until it is no longer argmax, transitioning to Case 1.

Any reliance to the non-causal factors (including non-causal content and environmental factors) corresponding to $\boldsymbol{x}^*$ is unstable and self-defeating, and will transition to ignorance (Case 1). Therefore, the model cannot converge to a minimum that relies on any specific non-causal factor to fit negative labels.
\end{proof}

\subsubsection{Discussion: Max-Pooling's Behavior}

This proof reveals that max-pooling-based methods are not affected by spurious correlations in negative bags, demonstrating strong robustness. The experiments on semi-synthetic datasets in the main text also provide strong empirical evidence.

\textbf{Intuitive Understanding of Model Behavior}: For instances containing factors that could serve as the signals for negative labels, their scores need to be very low, making them nearly impossible to become argmax. In fact, in WSI datasets, high-scoring instances in negative bags typically share some commonalities with positive instances, such as higher cellular density and more complex morphological structures. 

By driving down the scores of any concept observed in negative bags and identifying certain concepts that are absent in all negative bags but present in positive bags, the model learns to use these exclusive \textbf{positive concepts as evidence} to establish classification boundaries. The prediction score is based on \textbf{``whether this positive evidence appears''}, which aligns with the standard multiple instance learning assumption.

\subsection{Theorem 1: Environmental Robustness}

\begin{theorem}[Environmental Robustness]
\label{thm:environmental_robustness}
Under Assumptions A1--A4, for data generated according to the process defined in Section 3.2.1, any global minimizer $\theta^*$ of the empirical risk $\mathcal{R}_n(\theta)$ for a max-pooling-based model is environment-invariant, i.e., its scoring function depends only on $\boldsymbol{z}_c$.
\end{theorem}

\begin{proof}
\textbf{Step 1: Environmental factors in negative bags do not affect prediction.}

By Lemma~\ref{lem:unlearnability}, the model cannot use environmental factors in negative bags as evidence for negative labels. Furthermore, by Lemma~\ref{lem:opt_direction} (Optimization Direction), the model also cannot use them as evidence for positive labels.

\textbf{Step 2: Factors in positive bags cannot signal negative labels.}

By Lemma~\ref{lem:opt_direction}, the model cannot learn to use any factor in positive bags as evidence for negative labels.

\textbf{Step 3: Environmental factors in positive bags cannot signal positive labels.}

Suppose, for contradiction, that a classifier at a global minimum $\theta^*$ relies on some environmental factor $\boldsymbol{z}_e' \in \mathcal{E}$ as evidence for positive labels. That is, the model assigns high scores to instances $\boldsymbol{x}$ with $\boldsymbol{z}_e = \boldsymbol{z}_e'$ to fit positive bag labels.

By Assumption A4 (Multi-Environment Co-occurrence), there exists a negative bag $B_k \in \mathcal{B}^-$ that also contains an instance $\boldsymbol{x}^*$ generated from $\boldsymbol{z}_e'$.

Since the model assigns high scores based on $\boldsymbol{z}_e'$, this instance will also receive a high score:
\begin{equation}
s_{\theta^*}(\boldsymbol{x}^*) > 0.
\end{equation}

By max-pooling, the bag-level score for the negative bag satisfies:
\begin{equation}
S_{\theta^*}(B_k) = \max_{\boldsymbol{x} \in B_k} s_{\theta^*}(\boldsymbol{x}) \geq s_{\theta^*}(\boldsymbol{x}^*) > 0.
\end{equation}

This leads to a positive loss on the negative bag (with $Y_k = 0$):
\begin{equation}
\mathcal{L}_k(\theta^*) = -\log(1 - S_{\theta^*}(B_k)) > 0.
\end{equation}

Therefore, the total empirical risk is:
\begin{equation}
\mathcal{R}_n(\theta^*) = \frac{1}{n}\sum_{i=1}^{n}\mathcal{L}_i(\theta^*) \geq \frac{1}{n}\mathcal{L}_k(\theta^*) > 0.
\end{equation}

This contradicts the assumption that $\theta^*$ is a global minimizer with $\mathcal{R}_n(\theta^*) = 0$.

Therefore, the model cannot rely on $\boldsymbol{z}_e'$ as a basis for prediction, and can only rely on $\boldsymbol{z}_c$ for prediction.
\end{proof}

\subsection{Theorem 2: Focusing on Causal Factors}

\begin{theorem}[Focusing on Causal Factors]
\label{thm:causal_focus}
Under Assumptions A1--A5, for data generated according to the process defined in Section 3.2.1, any global minimizer $\theta^*$ of the empirical risk $\mathcal{R}_n(\theta)$ for a max-pooling-based model yields an invariant \emph{causal classifier}. Its instance-level scoring function $s_{\theta^*}(\boldsymbol{x})$ depends solely on the causal factor $\boldsymbol{z}_{\text{causal}}$ for prediction, while ignoring environmental factors ($\mathcal{E}$) and non-causal content factors ($\{\boldsymbol{z}_{n_1}, \ldots, \boldsymbol{z}_{n_k}\}$).
\end{theorem}

\begin{proof}
Building on Theorem~\ref{thm:environmental_robustness}, we only need to prove that the model does not rely on negative content factors for prediction.

Suppose, for contradiction, that a classifier at a global minimum $\theta^*$ relies on some negative content factor $\boldsymbol{z}_c' \in \{\boldsymbol{z}_{n_1}, \ldots, \boldsymbol{z}_{n_k}\}$ as evidence for positive labels. That is, the model assigns high scores to instances $\boldsymbol{x}$ with $\boldsymbol{z}_c = \boldsymbol{z}_c'$ to fit positive bag labels.

By Assumption A5 (Negative Concept Co-occurrence), there exists a negative bag $B_k \in \mathcal{B}^-$ that also contains at least one instance $\boldsymbol{x}^*$ generated from $\boldsymbol{z}_c'$.

Since the model assigns high scores based on $\boldsymbol{z}_c'$, this instance will also receive a high score:
\begin{equation}
s_{\theta^*}(\boldsymbol{x}^*) > 0.
\end{equation}

By max-pooling, the bag-level score for the negative bag satisfies:
\begin{equation}
S_{\theta^*}(B_k) = \max_{\boldsymbol{x} \in B_k} s_{\theta^*}(\boldsymbol{x}) \geq s_{\theta^*}(\boldsymbol{x}^*) > 0.
\end{equation}

This leads to a positive loss on the negative bag (with $Y_k = 0$):
\begin{equation}
\mathcal{L}_k(\theta^*) = -\log(1 - S_{\theta^*}(B_k)) > 0.
\end{equation}

Therefore, the total empirical risk is:
\begin{equation}
\mathcal{R}_n(\theta^*) = \frac{1}{n}\sum_{i=1}^{n}\mathcal{L}_i(\theta^*) \geq \frac{1}{n}\mathcal{L}_k(\theta^*) > 0.
\end{equation}

This contradicts the assumption that $\theta^*$ is a global minimizer with $\mathcal{R}_n(\theta^*) = 0$.

Therefore, the model cannot rely on $\boldsymbol{z}_c'$ as a basis for prediction.

Theorem~\ref{thm:environmental_robustness} excludes dependence on environmental factors $\mathcal{E}$. The only remaining predictive pathway is through the causal factor $\boldsymbol{z}_{\text{causal}}$.
\end{proof}

\section{Discussion on Assumptions}
\label{sec:appendix_assumptions_discussion}
\subsection{Evidence for A4 and A5 in real datasets}
\label{subsec:assumption_violation}
\paragraph{Assumption A4 (multi-environment co-occurrence)}
For the Camelyon16 dataset, both training and test sets contain tumor-positive and normal slides from two institutions (Radboud UMC and UMC Utrecht), with each center contributing both metastasis and normal lymph node sections~\cite{bejnordi2017Camelyon16}. For Camelyon17, the data are collected from five centers, and each center similarly provides both positive and negative slides\cite{bejnordi2017diagnostic}. The PANDA prostate cancer dataset combines approximately 10,616 WSIs from Radboud UMC (Netherlands) and Karolinska Institute (Sweden), and each site provides both cancer-positive and benign samples \cite{bulten2022artificial}. Designing multi-center datasets, with each center covering both positive and negative cases, has become standard practice in computational pathology.

\paragraph{Assumption A5 (negative concept co-occurrence)} A5 requires that the various normal tissue types and background patterns that appear in tumor slides also appear in normal slides. In the Camelyon16 dataset, apart from the cancerous regions, positive slides typically also contain extensive normal tissue components such as lymphatic tissue, fat, etc~\cite{bejnordi2017Camelyon16}. For lung tumor slides, malignant regions are often surrounded by rich normal structures, including bronchial epithelium, alveolar spaces, inflammatory cells, and stromal cells~\cite{coudray2018classification}. In the PANDA dataset, even cancer-positive biopsy cores often contain benign glands and stromal areas\cite{bulten2022artificial}. These normal tissue structures are intrinsic parts of human anatomy and may also frequently present in normal slides.

\subsection{Discussion on Assumption Violations}
Although our theoretical assumptions (A4 and A5) hold in the majority of Whole Slide Image (WSI) analysis scenarios, we discuss extreme cases where these assumptions may be violated to evaluate the reliability boundaries of our algorithm.

\paragraph{Violation of Assumption A4}
Consider the case where there exists an environment factor $\boldsymbol{z}_e'$ that appears only in positive bags of the training set. In this situation, since no negative bag contains patches $\boldsymbol{x}'$ generated from $\boldsymbol{z}_e'$, the model never receives a penalty even if it relies on $\boldsymbol{z}_e'$ as evidence for the positive label. As a result, the model may exploit $\boldsymbol{z}_e'$, which appears only in positive bags, as a shortcut.

When $\boldsymbol{z}_e'$ appears only at a low frequency in a small number of positive slides, relying solely on $\boldsymbol{z}_e'$ is insufficient to fit the remaining positive slides. The model is still forced to learn the content factor $\boldsymbol{z}_c$ to fit the dataset. Therefore, max-pooling methods can tolerate mild violations of Assumption A4. In contrast, when $\boldsymbol{z}_e'$ appears abundantly in positive slides but is completely absent in negative slides, the model may learn to rely on $\boldsymbol{z}_e'$ as a signal for positive label.


However, attention-based methods also fail in this scenario. Moreover, consider a case consistent with A4 where some environmental factor 
$\boldsymbol{z}_e^{*}$ appears only in negative slides and never in positive slides. As discussed in  Lemma~\ref{lem:unlearnability}, the max-pooling methods will ignore such $\boldsymbol{z}_e^{*}$. In contrast, attention models can learn $\boldsymbol{z}_e^{*}$ as a strong negative signal.

\paragraph{Violation of Assumption A5}
Similarly, if there exists a negative content factor 
$\boldsymbol{z}_c' \in \{\boldsymbol{z}_{n_1}, \ldots, \boldsymbol{z}_{n_k}\}$ that appears only in positive bags, the model may also learn to treat it as a positive shortcut, failing to distinguish between $\boldsymbol{z}_{\text{causal}}$ and $\boldsymbol{z}_c'$. Similar to the discussion in Assumption A4 above, when $\boldsymbol{z}_c'$ appears only at a low frequency in a small number of positive slides, the model still needs to learn $\boldsymbol{z}_{\text{causal}}$ in order to fit all slides. The max-pooling models remain robust to mild violations of A5. If $\boldsymbol{z}_c'$ appears frequently in positive slides, the model may rely on both $\boldsymbol{z}_c'$ and $\boldsymbol{z}_{\text{causal}}$ as evidence for the positive label.

In this case, the attention model will also exhibit this issue. Furthermore, max-pooling is immune to patterns appearing only in negative slides, whereas attention models can exploit them as shortcuts.

In the real world, there exist some cases where certain patterns $\boldsymbol{z}_c'$ co-occur with cancer. For example, some tumor tissues may be accompanied by a fibrotic response, forming a dense desmoplastic stroma composed of fibroblasts and collagen \cite{masugi2022desmoplastic}. These tissues rarely appear in normal slides. We argue that as long as the distribution of such patterns is relatively stable in the training and test sets, reliance on such patterns does not affect the model's robustness and generalization performance.  Furthermore, some tissue structures that appear alongside tumor tissues are also commonly used by pathologists in clinical practice as auxiliary evidence for diagnosis~\cite{masugi2022desmoplastic,kim2019inter}.


\section{Evidence of Rote Memorization}
\label{sec:rote-memorization}
\begin{figure}
    \centering
    \includegraphics[width=1\linewidth]{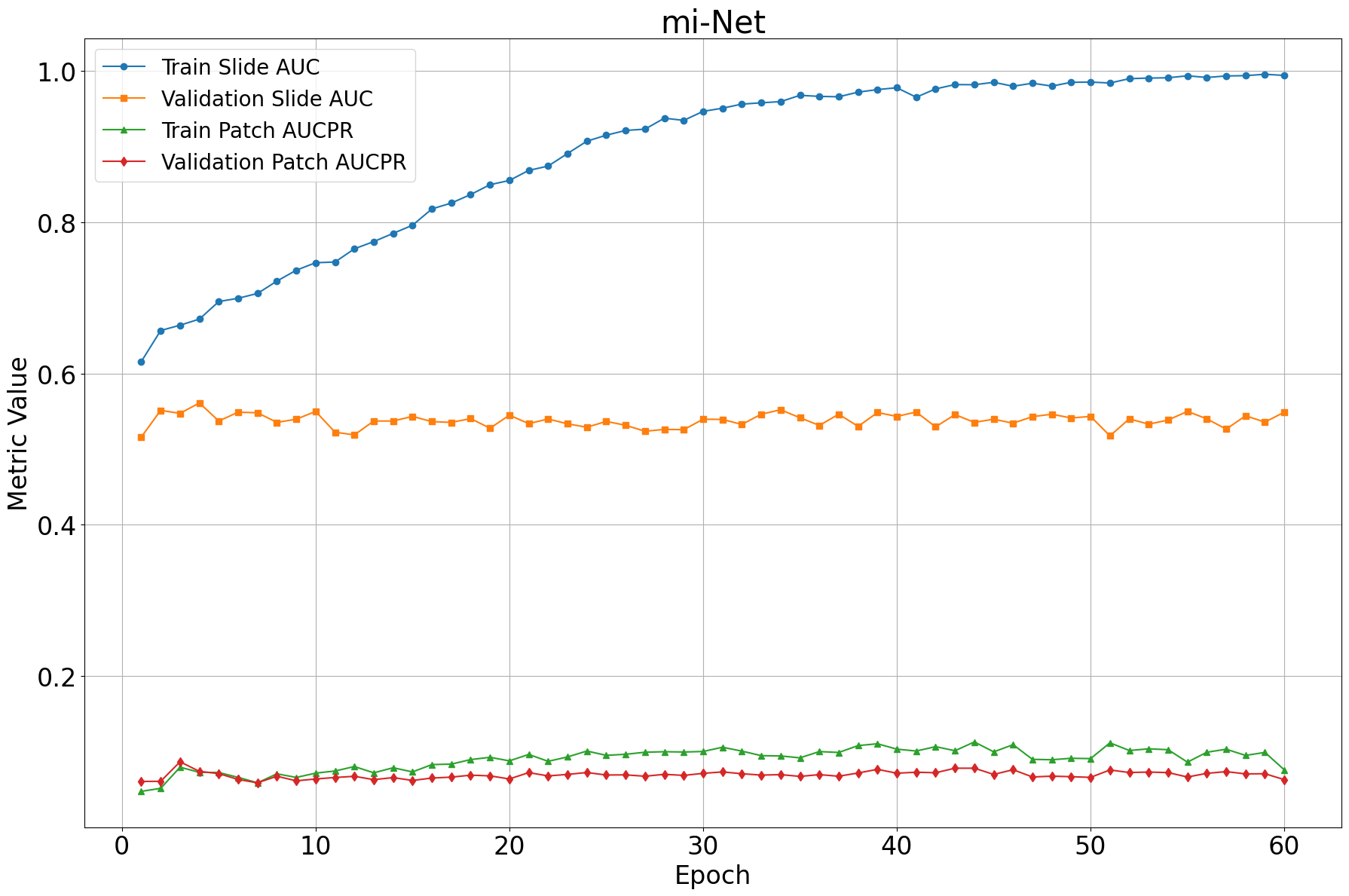}
    \caption{Training and validation metrics over epochs for
the mi-Net model on the Camelyon16 dataset.}
    \label{fig:minet_memor}
\end{figure}

As shown in Figure~\ref{fig:minet_memor}, mi-Net almost perfectly fits the training set at the slide-level, while it produces almost random predictions in the validation set and fails almost completely for the patch-level predictions. If the model had truly learned tumor patterns, the classifier should have achieved a respectable AUCPR at the patch level. These results suggests that the model relies on memorizing specific instance features as cues for bag labels, instead of learning meaningful patterns.

\section{Why Attention-Based MIL Methods Are Unstable} \label{sec:attention}

Attention-based methods aggregate weighted representations of all instances within a bag into a bag-level feature. This holistic aggregation strategy leads to gradients flowing through all attended instances during training. Consequently, all patterns including environmental factors $\boldsymbol{z}_e \in \mathcal{E}$ and content factors $\boldsymbol{z}_c \in \{\boldsymbol{z}_{causal}, \boldsymbol{z}_{n_1}, \ldots, \boldsymbol{z}_{n_k}\}$ can potentially be attended to as evidence for the presence of positive or negative labels.

In particular, patterns that appear only in negative bags can be learned as strong ``negative evidence,'' which directly conflicts with the standard MIL assumption that only the presence of positive instances should determine the bag label~\cite{dietterich1997solving,raff2023reproducibilityinMIL}.

Furthermore, this freedom to exploit environmental factors $\boldsymbol{z}_e$ also affects the performance of attention methods on hard positive instances. During training, attention models may overly focus on salient positive instances. In bags containing only a small number of hard positive instances, attention mechanisms can exploit $\boldsymbol{z}_e$ patterns to fit the label, thereby leading to missed diagnoses. Additionally, another scenario leading to missed diagnoses occurs when the model learns certain $\boldsymbol{z}_e$ as evidence for negative labels. When instances containing such $\boldsymbol{z}_e$ appear in positive bags, attending to these instances will decrease the bag's prediction score.

For instance-level prediction, attention methods may assign high scores to tumor-irrelevant regions based on environmental patterns~\cite{lin2023IBML,kaczmarzyk2024explainable}. Moreover, attention scores can only serve as a reference and do not reflect the degree of positiveness of individual instances. This is because attention scores merely reflect the contribution of each instance to the overall bag prediction~\cite{javed2022additive,kaczmarzyk2024explainable,jang2024MIL_learnable}.

\section{Additional Quantitative Experiments} \label{Additional Quantitative Experiments}
\begin{table}[htbp]
\centering
\caption{Ablation study on Camelyon16 Dataset. Dropout is set at 0.4 for the second model.}
\label{mi-Net}
\resizebox{\columnwidth}{!}{%
\begin{tabular}{lccc}
\toprule
\textbf{Model} & \textbf{Mini-batch} & \textbf{Slide AUC} & \textbf{Patch F1} \\
\midrule
mi-Net         & & 0.757 $\pm$ 0.122 & 0.298 $\pm$ 0.167 \\
mi-Net+dropout & & 0.801 $\pm$ 0.027 & 0.384 $\pm$ 0.033 \\
mi-Net+dropout     & \checkmark & 0.635 $\pm$ 0.133 & 0.248 $\pm$ 0.118 \\
FocusMIL(SGD)  &  & 0.814 $\pm$ 0.017 & 0.386 $\pm$ 0.022 \\
FocusMIL       & \checkmark & 0.871 $\pm$ 0.011 & 0.400 $\pm$ 0.011 \\
\bottomrule
\end{tabular}
}
\end{table}

\begin{table*}[htbp]
\centering
\caption{Ablation study on the KL divergence coefficient $\beta$. Results are reported as mean $\pm$ standard deviation of AUC across Camelyon16 (C16), Camelyon17 (C17), and TCGA-NSCLC datasets.}
\label{tab:beta-ablation}
\begin{tabular}{lccc}
\toprule
$\beta$ & AUC(C16) & AUC(C17) & AUC(TCGA) \\
\midrule
0                & $0.7130 \pm 0.155$ & $0.5121 \pm 0.028$ & $0.9311 \pm 0.005$ \\
$1\times10^{-6}$ & $0.7574 \pm 0.122$ & $0.6725 \pm 0.123$ & $0.9336 \pm 0.005$ \\
$1\times10^{-4}$ & $0.8667 \pm 0.014$ & $0.8695 \pm 0.020$ & $0.9320 \pm 0.002$ \\
$1\times10^{-3}$ & $0.8701 \pm 0.011$ & $0.8771 \pm 0.019$ & $0.9350 \pm 0.003$ \\
$1\times10^{-2}$ & $0.8610 \pm 0.013$ & $0.8719 \pm 0.008$ & $0.9357 \pm 0.002$               \\
$1\times10^{-1}$ & $0.8520 \pm 0.012$ & $0.8613 \pm 0.006$ & $0.9336 \pm 0.004$                 \\
$1$              & $0.7687 \pm 0.009$ & $0.6280 \pm 0.072$ & $0.9151 \pm 0.010$                    \\
\bottomrule
\end{tabular}
\end{table*}

\begin{table*}[htbp]
\centering
\caption{Ablation study on batch size.}
\label{tab:batchsize-ablation}
\begin{tabular}{lccc}
\toprule
Batch size & AUC(C16) & AUC(C17) & AUC(TCGA) \\
\midrule
1 & $0.8158 \pm 0.025$ & $0.8584 \pm 0.019$ & $0.9313 \pm 0.005$ \\
3 & $0.8682 \pm 0.009$ & $0.8771 \pm 0.019$ & $0.9350 \pm 0.003$ \\
5 & $0.8706 \pm 0.018$ & $0.8706 \pm 0.012$ & $0.9325 \pm 0.003$ \\
7 & $0.8721 \pm 0.010$ & $0.8738 \pm 0.010$ & $0.9339 \pm 0.003$ \\
9 & $0.8512 \pm 0.010$ & $0.8519 \pm 0.006$ & $0.9317 \pm 0.002$ \\
\bottomrule
\end{tabular}
\end{table*}

\subsection{Ablation Study}
\subsubsection{Ablation Study on the VIB Module}
We first investigate the performance of the max-pooling model without the VIB module. As shown in Table~\ref{mi-Net}, the baseline model mi-Net (a feed-forward neural network with a classifier) exhibits very large variance in results. This is because the method sometimes memorizes instance features, failing to learn any meaningful patterns. When dropout is set above 0.4, this severe overfitting problem disappears. However, when training mi-Net using multi-slide mini-batch gradient descent, this ``training failure'' reappears. FocusMIL employs a Variational Information Bottleneck (VIB) to learn a compact yet expressive latent distribution, which helps prevent these failures and achieves a slice-level AUC that is 1.3\% higher than mi-Net with dropout 0.4.

Table~\ref{tab:beta-ablation} reports the results of different KL divergence coefficients $\beta$. When $\beta = 0$ or $10^{-6}$, the model still suffers from training failures on Camelyon16. On Camelyon17, the model is even more prone to such ``memorization'' failures. When $\beta > 0.1$, the performance degrades on three datasets. This is because an excessively large $\beta$ forces the model to learn an overly compressed latent representation, which in turn reduces its capacity to encode label-relevant information. 

It is worth noting that on the TCGA-NSCLC dataset, VIB has minimal effect. This is because the tumor regions in this dataset are large, and cancer cell features are highly distinctive. It's easy for the model to learn cancer patterns.

\subsubsection{Ablation Study on Multi-Slide Mini-Batch Gradient Descent}

Table~\ref{tab:batchsize-ablation} presents results across different batch sizes on the three datasets. For the Camelyon16 dataset, batch sizes between 3 and 7 significantly outperform batch size 1. For Camelyon17, batch sizes greater than 1 also show improvements. 


For TCGA-NSCLC, since its cancer regions exhibit highly distinctive features, a batch size of 1 is already sufficient. Furthermore, since FocusMIL does not rely on environment information $\bm{z}_e$ for classification, a higher slide AUC actually indicates better classification performance on hard instances, which is further supported by our visualization results.

\begin{table*}[htbp]
\centering
\caption{Performance of MIL methods using CTransPath features on Camelyon16 dataset.}
\label{tab:ctrans_camelyon}
\resizebox{0.85\textwidth}{!}{
\begin{tabular}{lcccc}
\toprule
\multirow{2}{*}{Method}
& \multicolumn{2}{c}{Slide-level} &
  \multicolumn{2}{c}{Patch-level} \\ \cmidrule(r){2-3} \cmidrule(r){4-5}
& AUC & ACC & AUCPR & F1-score \\
\midrule
ABMIL     & 0.9659 \scriptsize{(0.9624, 0.9694)} & 0.9375 \scriptsize{(0.9306, 0.9444)} & 0.3884 \scriptsize{(0.3335, 0.4433)} & 0.3402 \scriptsize{(0.2497, 0.4306)} \\
DSMIL     & 0.9299 \scriptsize{(0.9048, 0.9551)} & 0.9094 \scriptsize{(0.8931, 0.9256)} & 0.6639 \scriptsize{(0.6587, 0.6690)} & 0.6086 \scriptsize{(0.5986, 0.6186)} \\
TransMIL  & 0.9603 \scriptsize{(0.9357, 0.9849)} & 0.9281 \scriptsize{(0.8935, 0.9627)} & -- & -- \\
DTFD-MIL  & \underline{0.9739} \scriptsize{(0.9722, 0.9756)} & 0.9515 \scriptsize{(0.9472, 0.9559)} & 0.4535 \scriptsize{(0.4315, 0.4756)} & 0.3350 \scriptsize{(0.3072, 0.3628)} \\
IBMIL     & 0.9716 \scriptsize{(0.9691, 0.9740)} & 0.9562 \scriptsize{(0.9476, 0.9649)} & 0.4285 \scriptsize{(0.3977, 0.4594)} & 0.2927 \scriptsize{(0.2607, 0.3247)} \\
CATTMIL   & 0.9418 \scriptsize{(0.9270, 0.9566)} & 0.9281 \scriptsize{(0.9200, 0.9362)} & 0.6516 \scriptsize{(0.6247, 0.6785)} & 0.5979 \scriptsize{(0.5773, 0.6185)} \\
AEM       & 0.9513 \scriptsize{(0.9410, 0.9615)} & 0.9109 \scriptsize{(0.8920, 0.9298)} & 0.3554 \scriptsize{(0.2766, 0.4343)} & 0.2989 \scriptsize{(0.2173, 0.3805)} \\
Conjunctive & 0.9661 \scriptsize{(0.9615, 0.9707)} & 0.9500 \scriptsize{(0.9423, 0.9577)} & \underline{0.6727} \scriptsize{(0.6444, 0.7009)} & \underline{0.6107} \scriptsize{(0.5784, 0.6430)} \\
mi-Net    & \textbf{0.9757} \scriptsize{(0.9723, 0.9791)} & \underline{0.9593} \scriptsize{(0.9550, 0.9637)} & 0.6156 \scriptsize{(0.6007, 0.6305)} & 0.5793 \scriptsize{(0.5542, 0.6045)} \\
CausalMIL & 0.9700 \scriptsize{(0.9661, 0.9740)} & 0.9531 \scriptsize{(0.9412, 0.9650)} & 0.6640 \scriptsize{(0.6509, 0.6770)} & \underline{0.6107} \scriptsize{(0.5762, 0.6452)} \\
FocusMIL  & 0.9731 \scriptsize{(0.9725, 0.9738)} & \textbf{0.9609} \scriptsize{(0.9609, 0.9609)} & \textbf{0.6902} \scriptsize{(0.6821, 0.6983)} & \textbf{0.6893} \scriptsize{(0.6772, 0.7014)} \\
\bottomrule
\end{tabular}}
\end{table*}

\begin{table*}[htbp]
\centering
\caption{Performance of MIL methods on Camelyon17 and TCGA-NSCLC.}
\label{tab:cam17_tcga_advanced}
\resizebox{0.85\textwidth}{!}{
\begin{tabular}{lcccc}
\toprule
\multirow{2}{*}{Method} & \multicolumn{2}{c}{Camelyon17} & \multicolumn{2}{c}{TCGA-NSCLC} \\
\cmidrule(lr){2-3}\cmidrule(lr){4-5}
& AUC & F1-score & AUC & ACC \\
\midrule
ABMIL     & 0.9264 \scriptsize{(0.9156, 0.9372)} & 0.8909 \scriptsize{(0.8880, 0.8938)} & 0.9686 \scriptsize{(0.9673, 0.9699)} & 0.9209 \scriptsize{(0.9177, 0.9242)} \\
DSMIL     & 0.9212 \scriptsize{(0.9035, 0.9390)} & 0.8904 \scriptsize{(0.8851, 0.8957)} & 0.9643 \scriptsize{(0.9605, 0.9680)} & 0.8914 \scriptsize{(0.8888, 0.8940)} \\
TransMIL  & 0.9387 \scriptsize{(0.9274, 0.9500)} & \underline{0.8932} \scriptsize{(0.8871, 0.8993)} & \textbf{0.9822} \scriptsize{(0.9814, 0.9831)} & \underline{0.9324} \scriptsize{(0.9259, 0.9388)} \\
DTFD-MIL  & 0.9447 \scriptsize{(0.9341, 0.9553)} & 0.8870 \scriptsize{(0.8788, 0.8952)} & \underline{0.9809} \scriptsize{(0.9795, 0.9823)} & 0.9286 \scriptsize{(0.9213, 0.9358)} \\
IBMIL     & 0.9410 \scriptsize{(0.9308, 0.9512)} & 0.8854 \scriptsize{(0.8764, 0.8944)} & 0.9790 \scriptsize{(0.9775, 0.9805)} & 0.9257 \scriptsize{(0.9225, 0.9358)} \\
CATTMIL   & 0.9325 \scriptsize{(0.9192, 0.9458)} & 0.8912 \scriptsize{(0.8869, 0.8955)} & 0.9685 \scriptsize{(0.9649, 0.9721)} & 0.9011 \scriptsize{(0.8979, 0.9043)} \\
AEM       & 0.9376 \scriptsize{(0.9328, 0.9423)} & 0.8851 \scriptsize{(0.8777, 0.8925)} & 0.9722 \scriptsize{(0.9700, 0.9744)} & 0.9236 \scriptsize{(0.9201, 0.9271)} \\
Conjunctive & 0.9353 \scriptsize{(0.9136, 0.9570)} & 0.8919 \scriptsize{(0.8827, 0.9011)} & 0.9805 \scriptsize{(0.9795, 0.9816)} & 0.9248 \scriptsize{(0.9134, 0.9361)} \\
mi-Net    & 0.9442 \scriptsize{(0.9307, 0.9577)} & 0.8882 \scriptsize{(0.8819, 0.8946)} & 0.9683 \scriptsize{(0.9635, 0.9703)} & 0.9267 \scriptsize{(0.9199, 0.9334)} \\
CausalMIL & \textbf{0.9537} \scriptsize{(0.9445, 0.9628)} & 0.8885 \scriptsize{(0.8799, 0.8970)} & 0.9725 \scriptsize{(0.9691, 0.9758)} & 0.9228 \scriptsize{(0.9141, 0.9316)} \\
FocusMIL  & \underline{0.9529} \scriptsize{(0.9456, 0.9602)} & \textbf{0.8974} \scriptsize{(0.8875, 0.9073)} & 0.9744 \scriptsize{(0.9717, 0.9771)} & \textbf{0.9352} \scriptsize{(0.9320, 0.9388)} \\
\bottomrule
\end{tabular}}
\end{table*}

\subsubsection{Additional Results Using Advanced Feature Extractors}
\textbf{Results on Camelyon16} As shown in Table~\ref{tab:ctrans_camelyon}, when using features extracted by CTransPath~\cite{wang2022ctranspath}, all models show a noticeable performance boost. In this case, the max-pooling-based methods still achieve significantly better results at the patch level. Thanks to the robust latent representations learned through VIB, FocusMIL achieves the best patch-level performance among all methods. For slide-level, three max-pooling-based methods achieve slightly leading results. Notably, DSMIL achieves better results compared to other attention-based methods. 
This improvement is attributed to DSMIL's first stream using a max-pooling-based instance classifier.Since CATTMIL employs DSMIL as its backbone, it also delivers competitive patch-level performance. Conjunctive-MIL is a hybrid model that combines instance-level predictions with an attention mechanism, and it likewise achieves strong patch-level results.

\textbf{Results on Camelyon17}
We directly use the features extracted by PathGen-CLIP from the public repository of the AEM paper~\cite{zhang2024aem}. As shown in Table~\ref{tab:cam17_tcga_advanced}, the three max-pooling-based methods still exhibit slightly better performance. We observe that attention-based methods also achieve strong performance.
This is because the feature extractor was pretrained with self-supervised contrastive learning and data augmentation.
Data augmentation can be seen as a form of intervention which can help the model learn representations that are more robust to bias~\cite{von2021self}. Nevertheless, max-pooling serves as an \textbf{additional reliable safeguard}.

\textbf{Results on TCGA-NSCLC}
We use features extracted by CTransPath. As shown in Table~\ref{tab:cam17_tcga}, we observe that FocusMIL achieves competitive results compared to other methods. When the training and test distributions are similar, max-pooling-based methods may not show a clear advantage.
The main strengths of FocusMIL lie in its robustness to domain shifts and its interpretability at the instance-level prediction.

\section{Computational Cost Analysis} \label{Computational Cost Analysis}
As shown in Table~\ref{tab:6}, FocusMIL has significantly fewer parameters compared to attention-based methods. Due to the use of mini-batch gradient descent with a batch size of 3, both CausalMIL and FocusMIL achieve superior training speeds. However, CausalMIL requires additional computations for reconstructing loss and the KL divergence loss of the conditional prior distribution, which results in slightly longer training times compared to FocusMIL.

\begin{table}[htbp]
\centering
\caption{Comparison of model size and training time per epoch of different MIL methods. The input dimension is set the same for each model to ensure fairness of comparison.}
\label{tab:6}
\resizebox{\columnwidth}{!}{%
\begin{tabular}{lcccc}
\toprule
\textbf{Model} & \textbf{Para.} & \textbf{Time} & \textbf{C16 AUC} & \textbf{C17 AUC} \\
\midrule
ABMIL     & 657k       & 3.35s    & 0.8052 & 0.7877 \\
DSMIL     & 856k       & 4.92s    & 0.7733 & 0.7069 \\
TransMIL  & 2.66M      & 9.56s    & 0.8352 & 0.6899 \\
DTFD-MIL  & 987k       & 6.02s    & 0.8619 & 0.8192 \\
CausalMIL & \underline{302k} & \underline{3.28s} & 0.8092 & 0.5104 \\
FocusMIL     & \textbf{167k}   & \textbf{3.23s}   & \textbf{0.8706} & \textbf{0.8719} \\
\bottomrule
\end{tabular}
}
\end{table}
\begin{figure*}
    \centering
    \includegraphics[width=0.75\linewidth]{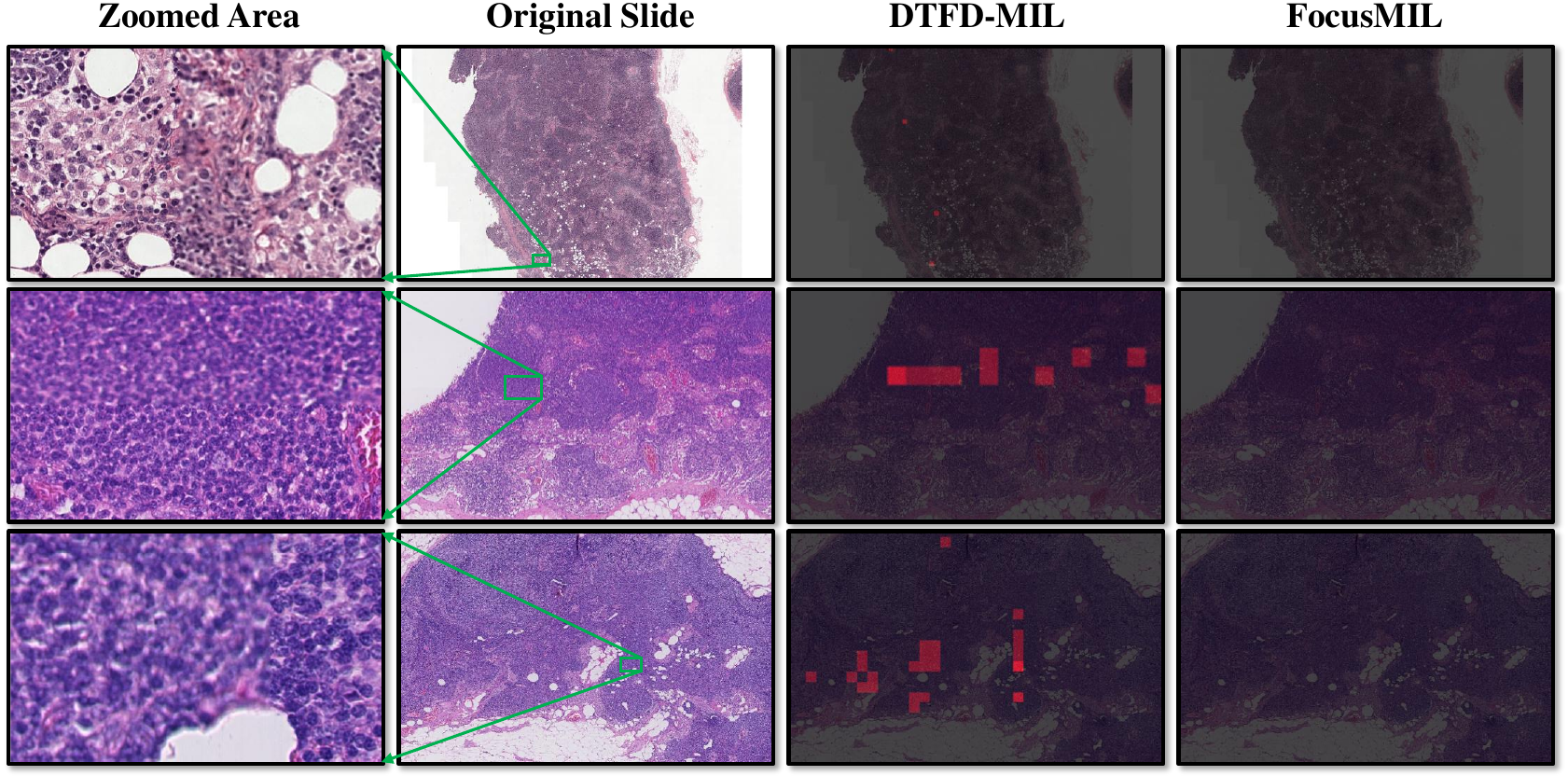}
    \caption{Visualization results of FocusMIL and DTFD-MIL on 3 normal slides. In the zoomed-in areas, some regions are relatively clear, while the rest are very blurred. DTFD-MIL may assign a high positive probability to the blurred regions.}
    \label{fig:normal}
\end{figure*}

\begin{figure*}
    \centering
    \includegraphics[width=0.75\linewidth]{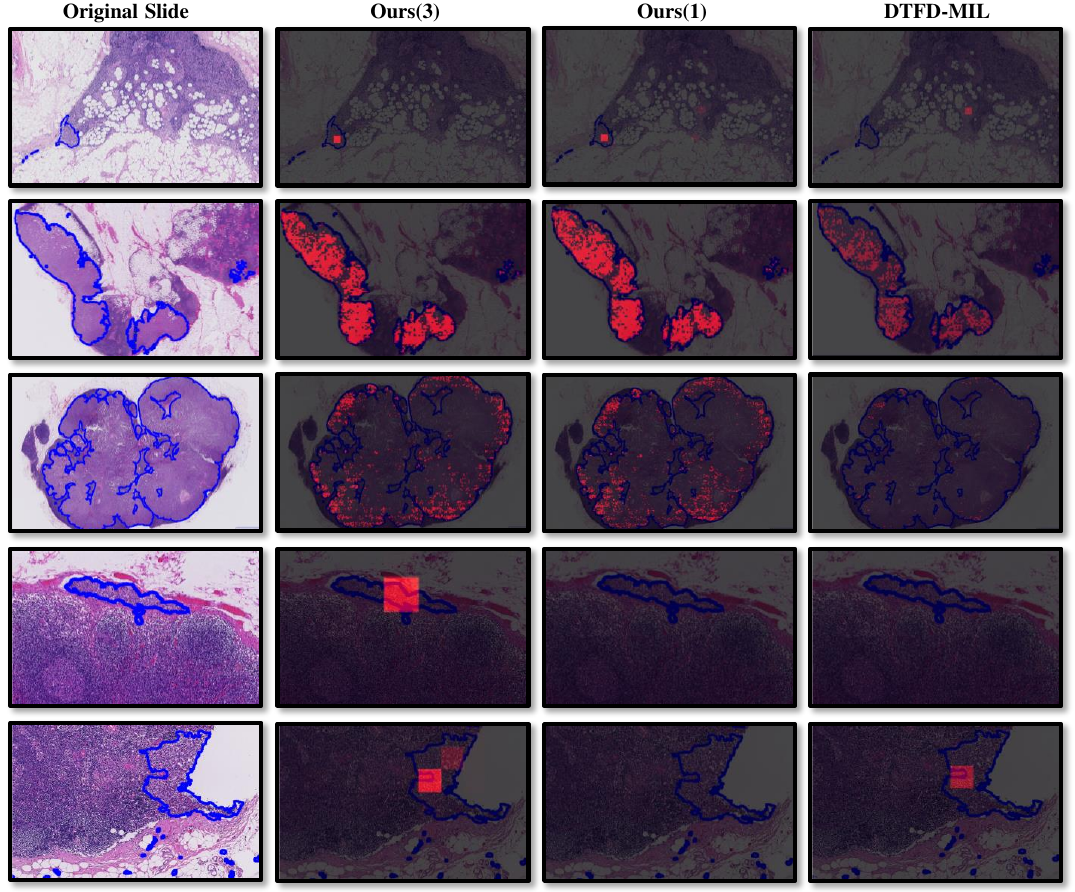}
    \caption{Additional visualization results of FocusMIL and DTFD-MIL on Camelyon16 dataset. Ours(3) and Ours(1) refer to the FocusMIL models trained with a batch size of 3 and 1, respectively. The tumour regions are delineated by the blue lines.}
    \label{fig:add visual}
\end{figure*}
\section{Additional Visualization Results} \label{additionVisual}
Notably, in the fourth slide of Figure 3 in the main text and the normal slides in Figure~\ref{fig:normal}, DTFD-MIL assigns high scores to many patches unrelated to the tumour. After zooming in on these patches, we observe that the tissues in these regions are significantly blurrier, which is associated with the slide preparation process. This indicates that DTFD-MIL relies on enviroment factor $\bm{z}_e$ for predictions. FocusMIL does not have any false-positive predictions
for these normal slides.

Figure~\ref{fig:add visual} demonstrates the visualization results of FocusMIL and DTFD-MIL for five tumour slides. FocusMIL with a training batch size of 3 perfectly predicts all tumour regions. When the training batch size was changed to 1, FocusMIL misses some small tumour regions. This result confirms that training with mini-batch gradient descent does help to establish better classification boundaries for salient and hard instances. For DTFD-MIL, its predictions for both large and small tumour areas are suboptimal due to the fact that it can overfit on salient instances.

\section{Experimental setup}  \label{Experi}
\subsection{Dataset and Metric}

\textbf{Camelyon16 dataset} is widely used for metastasis detection in breast cancer~\cite{bejnordi2017Camelyon16}. The dataset consists of 270 training and 129 testing WSIs, which yield roughly 2.7 million patches at 10× magnification (the second level in multi-resolution pyramid). Pixel-level labels are available for tumour slides. Every WSI is cropped into $256 \times 256$ patches without overlap, and background patches are discarded. A patch is labeled positive if it contains $25\%$ or more cancer areas. The numbers of tumorous versus normal patches are imbalanced as the positive patch ratios of positive slides in the training and testing sets of Camelyon16 are approximately 8.8\% and 12.7\%, respectively. We used random seeds to re-split the official training set into training and validation sets at an 8:2 ratio. The best model was selected on the validation set and finally tested on the official test set. The complete code for pre-processing will be made publicly available.

\textbf{Camelyon17 dataset} consists of 1,000 WSIs from five hospitals, categorized into different slide labels such as Normal, Isolated Tumour Cells, Micro-metastases, and Macrometastases~\cite{bejnordi2017diagnostic}. Since this study focuses on a binary classification task, we relabeled the slides into two categories: Normal and Tumour. Due to the absence of labels in the test set, we only used the 500 slides from the training set. Following the setup of AEM~\cite{zhang2024aem}, to evaluate the model's OOD generalization performance, we designated 200 slides from the fourth and fifth hospitals as the test set. The remaining 300 slides were split into training and validation sets with an 8:2 ratio.

\textbf{TCGA-NSCLC} includes two subtypes of lung cancer: Lung Adenocarcinoma (LUAD) and Lung Squamous Cell Carcinoma (LUSC), with a total of 1,054 diagnostic digital slides. Only slide-level labels are available for this dataset. We directly used the patches released by Li \textit{et al.}~\cite{DSMIL}.

\textbf{Evaluation Metrics
} 
Due to the high level of class imbalance at the patch level, we report Area Under the Precision-Recall Curve (AUCPR) and F1 score for evaluation. For slide-level classification on Camelyon16 dataset, we report AUC and Accuracy as the tumour and normal classes are balanced. We report AUC and F1-score on Camelyon17 dataset, as the classes are relatively imbalanced. On Camelyon16, we also evaluate tumour region localization performance by reporting free response operating characteristic curves (FROC)~\cite{bejnordi2017Camelyon16}.\\
\subsection{Implementation details}
 The encoder of FocusMIL consists of a neural network with one hidden layer, with ReLU as the activation function.
 We use AdamW optimizer with an initial learning rate of $1 \times 10^{-4}$ to update the model weights during the training. The dimension of the latent factor is set to 35. $\beta$ is set to 0.001. The mini-batch size for training FocusMIL model is 3. All experiments are conducted 5 times, and the mean and 0.95 confidence intervals (CI) are reported.

For max-pooling-based methods, we directly use the classifier output as the patch-level prediction. For ABMIL, following the DTFD paper~\cite{zhang2022dtfd}, we use the normalized attention scores for patch-level prediction. For DTFD-MIL, we use the instance probability derivation proposed in their paper for patch-level prediction.
 
For the attention-based methods, we build other models based on their officially released codes and conduct grid searches for key hyperparameters. For IBMIL, we employ DTFD-MIL as the backbone model~\cite{lin2023IBML}. For CATTMIL, we employ DSMIL as the backbone model~\cite{wu2024causal}. For ABMIL, we use gated attention with dropout rate of 0.25 to achieve optimal performance. For CausalMIL~\cite{zhang2022CausalMIL}, both its encoder and decoder are set up as neural networks with hidden layer neurons 128 and two other fully-connected layers are used to carve factorized prior distribution conditioned on the bag information. All the experiments are conducted with a single Nvidia RTX4090 GPU. The code will be made publicly available upon acceptance.

For all experiments, we perform dataset splits using five random seeds. All models use identical splits to ensure fair comparison. We made our best efforts to ensure optimal performance for all baselines.

To generate the visualization results, the models are not deliberately selected for fairness of comparison. Both FocusMIL with a training batch size of 3 and DTFD-MIL models achieved a test
slide AUC of approximately 0.87.

\end{document}